\documentclass{article}

\usepackage[final]{corl_2020} 

\usepackage{enumitem}
\usepackage{amsmath} 
\usepackage{mathtools}
\usepackage{algorithm} 
\usepackage{algpseudocode}
\usepackage{graphicx}
\usepackage{subcaption}
\usepackage{wrapfig}
\usepackage{comment}
\usepackage{subfiles}
\usepackage[resetlabels]{multibib}
\usepackage[belowskip=-6pt,aboveskip=6pt]{caption}

\newcites{S}{References}

\include{pythonlisting}

\definecolor{dgreen}{rgb}{0.0,0.545,0.0}
\definecolor{dblue}{rgb}{0.0, 0.7, 1.0}
\definecolor{dpurple}{rgb}{0.5, 0.0, 0.5}

\newcommand{\mx}{\mathbf{x}}
\newcommand{\muu}{\mathbf{u}}

\title{Soft Multicopter Control Using \\ Neural Dynamics Identification}

\author{
  Yitong Deng${^{1*}}$, Yaorui Zhang$^1$, Xingzhe He$^1$, Shuqi Yang$^1$, Yunjin Tong$^1$\\
  \textbf{Michael Zhang$^2$, Daniel DiPietro$^1$, Bo Zhu$^1$}\\
  $^1$Dartmouth College, Computer Science Department \\$^2$Lawrenceville School\\
  \texttt{$^*$ yitong.deng.gr@dartmouth.edu}
}

\begin{document}
\maketitle


\begin{abstract}
We propose a data-driven method to automatically generate feedback controllers for soft multicopters featuring deformable materials, non-conventional geometries, and asymmetric rotor layouts, to deliver compliant deformation and agile locomotion. Our approach coordinates two sub-systems: a physics-inspired network ensemble that simulates the soft drone dynamics and a custom LQR control loop enhanced by a novel online-relinearization scheme to control the neural dynamics. Harnessing the insights from deformation mechanics, we design a decomposed state formulation whose modularity and compactness facilitate the dynamics learning while its measurability readies it for real-world adaptation. Our method is painless to implement, and requires only conventional, low-cost gadgets for fabrication. In a high-fidelity simulation environment, we demonstrate the efficacy of our approach by controlling a variety of customized soft multicopters to perform hovering, target reaching, velocity tracking, and active deformation.
\end{abstract}

\hspace{36pt}\textbf{Video: } https://youtu.be/xYWp5jzOwkc


\section{Introduction}
Making a drone's body soft opens up brand new horizons to advance its maneuverability, safety, and functionalities.   
The intrinsic property of soft materials to deform and absorb energy during collision allows safe human-machine interactions \citep{rus2015design,lee2017soft,7551190}.
In circumscribed environments, soft drones can naturally deform their bodies to travel through gaps and holes, making them effective for emergency rescues.
Moreover, the ability to perform controlled deformation enables soft drones to perform secondary functionalities apart from aerial locomotion, such as flapping wings, grasping objects, and even operating machines, any additional mechanical parts. 

Despite the various advantages that this promises, to date a reliable and practical algorithm that controls soft drones to fly and deform has been lacking, due to multifaceted challenges. Unlike how it is for rigid drones, which are fully defined by 12-dimensional state vectors, describing the state of soft drones is far from trivial. Since a continuum body deforms in infinite DOFs, one needs to design discrete representations both compact ------ so that the underactuated control problem is feasible, and measurable ------ so that the controller can adjust to unmodelled errors. Even if such discretizations are obtained, the dynamic interplay of these state variables cannot be derived analytically in closed-form, as the dynamics in the full space is governed by complex PDEs. Finally, if one is to adopt machine-learning methods to model the dynamics, a problem is posed by the complexity of soft body simulation, which inevitably leads to scarce data, making brute-force learning an ill-fated avenue.

Bridging the deformation mechanics, deep learning, and optimal control, our method is so designed to successfully overcome the abovementioned challenges. First, we adopt the polar decomposition theorem to build a low-dimensional decomposed state space consisting of three geometric, latent variables each representing rotation, translation, and deformation, which can be synthesized from the readings of a set of onboard Inertial Measurement Units (IMUs). We define the interdependencies of these three variables and learn them with lightweight neural networks, thereby learning a neural simulator in a latent space on which the controller will be based. With automatic differentiation, we then extract the numeric gradients of the learned system to be controlled with a Linear Quadratic Regulator (LQR). Due to the fact that LQR requires the system to be linearized around fixed points that are inaccessible, we extend it with a novel online-relinearization scheme that iteratively converges to the desired target, bringing robustness and convenience for human piloting. 

\begin{figure}[t]
    \includegraphics[width = 1.\textwidth]{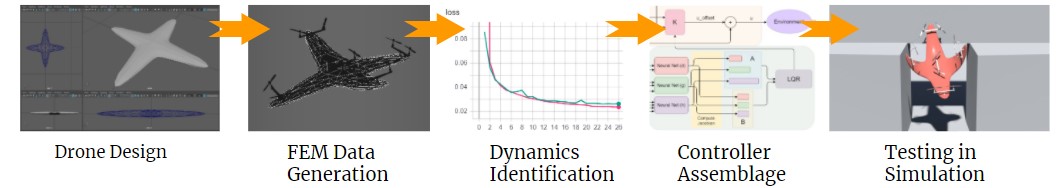}
    \\
    \caption{The 5-stage pipeline of our contorller generation system }\label{fig:ppline}
\end{figure}

As shown in Figure. \ref{fig:ppline}, our system takes soft drone geometries as input, and returns functionals that compute control matrices based on the drone's current state.
To the best of our knowledge, the proposed approach is the first to control soft drones that are meant to deform significantly in flight; we show that we can not only regulate such deformation for balanced locomotion, but also capitalize on the deforming ability to perform various feats in the air.

\section{Related Work}
\label{sec:previous works}
\textbf{Multicopter}\ \ \ \ \ Over the last few years, quadcopters have virtually dominated the commercial UAV industry, thanks to their simple mechanical structures, optimized efficiency for hovering, and easy-to-control dynamics that has been extensively studied by \citep{russnote,hoffmann2007quadrotor} 
and many more. 
Various methods have been successfully developed to control quadcopters, including PD/PID \citep{tayebi2004attitude}, LQR \citep{du2016computational,bouabdallah2004pid}, differential flatness \citep{mellinger2011minimum}, sliding mode \citep{waslander2005multi}, and MPC \citep{wang2015mpc} methods. 
Recent research explores the control of non-conventional multicopter designs, including ones with extra rotors \citep{baranek2012modelling}, asymmetric structures \citep{du2016computational}, articulated linkage \citep{zhao2018dragondrone}, or with gliding wings \citep{xu2019learning}. The problem of controlling drones fabricated with soft materials is still understudied.

\textbf{Aerial Deformation}\ \ \ \ \ Recent works have explored the potential of drones to actively deform in flight \citep{floreano2017foldable}. \citep{zhao2018dragondrone, anzai2018aerialdeform} achieve impressive results with their multi-linked drones in passing through small openings or grasping objects, but the added mechanical components in their designs imply additional cost, fabrication complexity, energy consumption, and maladroitness. \citep{falanga2018foldable} proposes a lightweight, planar folding mechanism actuated by servo motors that is effective in controlling the drones to travel through confined spaces, but the simplified mechanism limits the ability to perform extra functionalities such as grasping. \citep{fishman2020control} propose the incorporation of a cable-actuated soft gripper with a rigid quadcopter to achieve load manipulation. While previous works add additional actuators to control deformation, we control deformation jointly with locomotion using rotors only.
\textbf{Data-driven Soft Robot Control}\ \ \ \ \
The modeling and control of soft robots is a challenging problem due to the high DOFs and the non-linear dynamics \cite{rus2015design}, which together make closed-form solutions unfeasible to be derived \cite{spielberg2019learning}, and therefore making data-driven approaches favorable. \citep{basedreinforce1, reinforce2, freereinforce3}, and many more, use deep reinforcement learning to train neural network controllers for soft robots. \citep{neuraltraditional1, neuraltraditional2} marries machine learning with control theory, applying MPC or PD control methods on models learned from data. \cite{spielberg2019learning} propose the possibility of end-to-end supervised learning with a differentiable soft-body simulator, although their current design does not facilitate feedback mechanism in real-life due to the assumption of full state measurement.

\section{Methodology}
\label{sec:method}
\begin{wrapfigure}{r}{.5\textwidth}
 	\vspace{-0.5\baselineskip}
	\centering
 \includegraphics[width=0.5\textwidth]{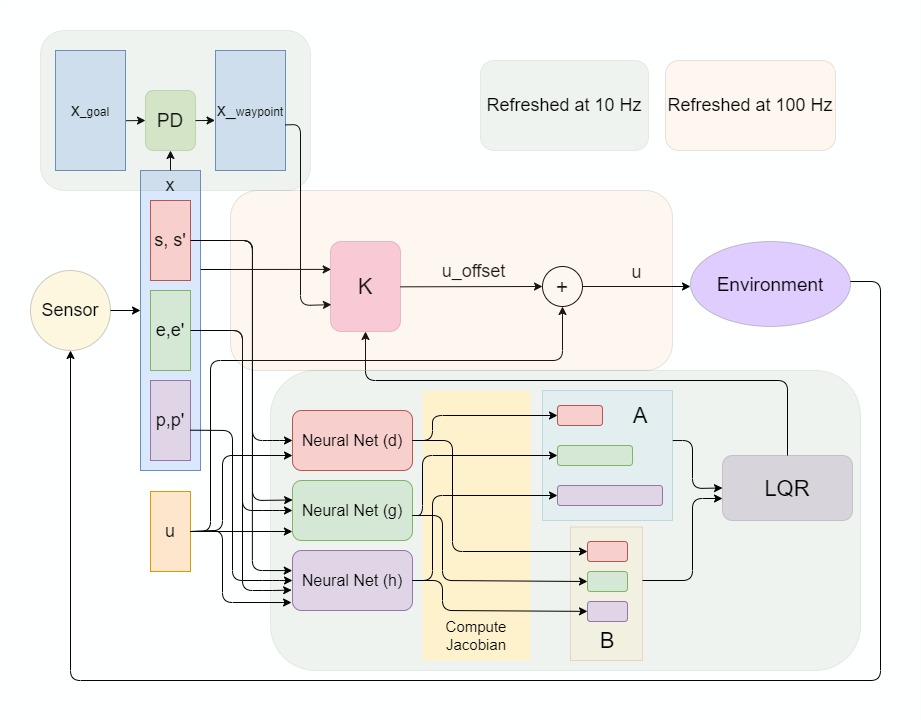}
    \\
    \caption{Method overview}\label{fig:overview}
\end{wrapfigure}

\textbf{Overview}\ \ \ \ \ The workflow of our proposed system is given in Figure. \ref{fig:overview}. We start with the IMU sensor readings, which are position, orientation and their rates of change at multiple locations of the drone's body, which will be processed to form the current state vector $\mx$ as the concatenation of the three decomposed latent variables $\mathbf{s}$, $\mathbf{e}$, and $\mathbf{p}$. 
This vector, 
along with the previously 
applied 
actuation 
$\muu$ 
will be passed into the three trained network modules as inputs. 
Then we extract the numeric gradients to form the Jacobian matrices, which will be used to assemble the matrices $\mathbf{A}$ and $\mathbf{B}$ representing the linearized dynamics. 
Then, $\mathbf{A}$ and $\mathbf{B}$ will be passed into the LQR algorithm to form the control matrix $\mathbf{K}$, given which we simply need to pass in our current state $\mx$, our next waypoint state $\mathbf{x}_{wp}$, and the current actuation $\muu$, to obtain the new actuation $\mathbf{u}_{next}$, which will be applied to the drone's rotors to complete the control loop. The $\mathbf{x}_{wp}$ will be computed by a Proportional-Derivative (PD) controller given final goal state $\mathbf{x}_{goal}$. The control loop will operate at $100Hz$ while the recomputation of $\mathbf{K}$ takes place at $10Hz$.

\textbf{Geometric State Decomposition}\ \ \ \ \ 
Let $\hat{\Omega}$, $\Omega$ denote the volume of the drone's undeformed and deformed geometry, and let $\hat{\mathbf{z}}$, $\mathbf{z}$ be their respective finite discretizations as particles forming the solid body, with $\mathbf{z} = \Phi(\hat{\mathbf{z}})$. We can always write:
\begin{equation}\label{eq:decomp}
    \mathbf{z}=\mathbf{R}S(\hat{\mathbf{z}})+\mathbf{p},
\end{equation}

for some rotation matrix $\mathbf{R}$, some position $\mathbf{p} \in \mathbf{R}^3$, and some non-linear function $S$ which reduces to $\Phi$ when $\mathbf{R = I}$ and $\mathbf{p = 0}$. In other words, we first define a local reference frame by the $SO3$ transformation resulting from $\mathbf{R}$ and $\mathbf{p}$, and express the deformation in such a frame with $S$, thereby decomposing the total transformation into a rigid component and a deformable component. In this way, the rigid, $SO3$ component can be simulated side by side with the deformable component $S$, an approach proposed by \citep{terzopoulos1988physically} and adopted by various works in the deformation simulation community \cite{pentland1989good, sorkine2007rigid, lu2016two}, taking advantage of the fact that the local-frame deformation $S$ is much nicer to work with than the world-frame deformation $\Phi$. In this work, we adopt the same philosophy and learn evolution rules for $\mathbf{R}$, $\mathbf{p}$, and $S$ in juxtaposition.

\begin{wrapfigure}{r}{.30\textwidth}
	\centering
	\vspace{-1.\baselineskip}
 \includegraphics[width=0.30\textwidth]{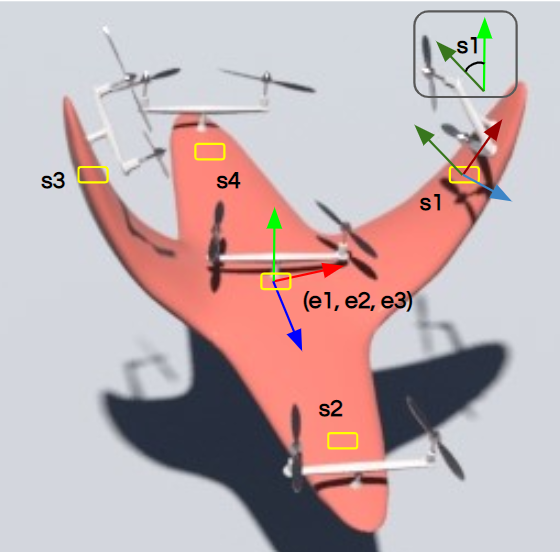}
    \\
    \caption{Sensor scheme}\label{fig:sep}
\end{wrapfigure}
\textbf{Measurability}
\ \ \ \ \ Since our goal is to build feedback controllers for $\mathbf{z}$, \textit{i.e}, $(\mathbf{R}, \mathbf{p}, S)$, we would need to define $\mathbf{R}$, $\mathbf{p}$, and $S$ in such a way that they can be measured by sensors. In this work, we adopt the following strategy. As shown in Figure. \ref{fig:sep} we plant Inertial Measuring Units (IMUs), which are compact, low-cost MEMS outputting rotational and translational status, at the drone's geometric center and around its periphery, next to the propellers. For the local frame rotation $\mathbf{R}$, we use the Euler angles $\mathbf{e} = (e1, e2, e3)$ measured at the geometric center. For local frame origin $\mathbf{p}$, we average the positional measurements of all the IMUs to approximate the center of mass. For the local frame deformation $S$, we obtain an approximate representation $\mathbf{s}$ by computing a scalar angle difference $s_i$ between the $Y$-axis measured by each peripheral IMU and the $Y$-axis defined by $\mathbf{R}$, and concatenate these scalars together. In the shown case, each entry of $\mathbf{s}$ represents how its associated wing is folded, where the sign indicates inward or outward and the magnitude indicates the angle. Such a design of $\mathbf{s}$ facilitates interactive piloting since it is intuitive to describe one's desired deformation in terms of the bent angles. Since we have parameterized $\mathbf{z}$ with $\mathbf{s}$, $\mathbf{e}$ and $\mathbf{p}$, controlling the dynamics 
of $\mathbf{z}$ now relegates to controlling the dynamics of $(\mathbf{s, e, p})$.

\textbf{Dynamic Coupling}\ \ \ \ \ The drone's body will be actuated by a set of $m$ propellers, each providing a scalar thrust along the normal of the surface it is planted on. Let $\mathbf{u} \in \mathbf{R}^m$ denote the thrusts. The second-order dynamics of $\mathbf{z}$ is given by the momentum conservation:

\begin{equation}\label{eq:dyn}
    m\ddot{\mathbf{z}} = f_{actuation}(\mathbf{z}, \mathbf{u}) + f_{stress}(\mathbf{z}) + f_{damping}(\dot{\mathbf{z}}) + m\mathbf{g}
\end{equation}
where $m$ represents the particle mass. We note here the translational and rotational symmetries of the experienced forces. Since $f_{actuation}$ depends on the rotors' normal directions which are computed locally with neighboring particles, it is unchanged when the whole of $\mathbf{z}$ is rotated or translated. The translational and rotational invariance of $f_{stress}$ and $f_{damping}$ is based on the principle of material frame-indifference in continuum mechanics, which states that the behavior of a material is independent of the reference frame \citep{speziale1998review}. 
Combining the symmetry with the equality $ \mathbf{z}=\mathbf{R}S(\hat{\mathbf{z}})+\mathbf{p}$, and ignoring the Coriolis forces resulting from the evolving $\mathbf{R}$, we get that:
\begin{equation}\label{eq:dyn2}
    m\ddot{\mathbf{z}} \approx \mathbf{R}f_{actuation}(S(\hat{\mathbf{z}}), \mathbf{u}) + \mathbf{R}f_{stress}(S(\hat{\mathbf{z}})) + \mathbf{R}f_{damping}(\dot{S(\hat{\mathbf{z}}})) + m\mathbf{g}.
\end{equation}
Since $\hat{\mathbf{z}}$ is constant, $S(\hat{\mathbf{z}})$ is a function of $\mathbf{s}$, and we know $\mathbf{R}$ is a function of $\mathbf{e}$, then the dynamics coupling of $\mathbf{s}$, $\mathbf{e}$, and $\mathbf{p}$ are as follows. Since $\ddot{\mathbf{p}}$ is the average of $\ddot{\mathbf{z}}$, it depends on $\mathbf{u}$, $\mathbf{e}$, $\dot{\mathbf{e}}$, $\mathbf{s}$ and $\dot{\mathbf{s}}$. Since $\mathbf{R}$ is measured in the local geometric center of $\mathbf{z}$, $\ddot{\mathbf{e}}$ depends on $\mathbf{u}$, $\mathbf{e}$, $\dot{\mathbf{e}}$, $\mathbf{s}$ and $\dot{\mathbf{s}}$ as well. For $\ddot{\mathbf{s}}$, since it is measured by projecting $\mathbf{z}$ onto the local frame by left-multiplying $\mathbf{R^{-1}}$, the $\mathbf{R}$ component in $\ddot{\mathbf{z}}$ cancels out for $\ddot{\mathbf{s}}$ and $\ddot{\mathbf{s}}$ depends on $\mathbf{u}$, $\mathbf{s}$ and $\dot{\mathbf{s}}$ only.

These interdependencies will be modelled by learned neural networks, in particular, we train networks $\{\mathbf{d}, \mathbf{g}, \mathbf{h}\}$ such that $\mathbf{\dot{s}}_{next}\mathbf{ =d(s,\dot{s}, u)}$, $\mathbf{\dot{e}}_{next} \mathbf{=g(s,\dot{s}, e,\dot{e}, u)}$, and $\mathbf{\dot{p}}_{next}\mathbf{=h(s,\dot{s}, e,\dot{e}, \dot{p}, u)}$.

\textbf{Learning the Dynamics} \ \ \ \ \
Training the networks $\mathbf{d}$, $\mathbf{g}$ and $\mathbf{h}$ can be done in a relatively straightforward fashion. The three networks share the same lightweight architecture consisting of four residual blocks \citep{resnet} featuring linear layers as previously explored by~\citep{weinanresnetode1,luresnetode2}. The three networks will be trained separately using the same reservoir of data samples of the form \{$\mathbf{s},\mathbf{e},\mathbf{p},\dot{\mathbf{s}},\dot{\mathbf{e}},\dot{\mathbf{p}},\mathbf{u},\dot{\mathbf{s}}_{next},\dot{\mathbf{e}}_{next},\dot{\mathbf{p}}_{next}$\} generated from Finite Element Method (FEM) simulation. We use Adam optimizer and L1 loss for optimization, with hyperparameter details given in the supplement. There are two techniques that we adopt that are worth reporting. First, for data generation, we would apply a constant random thrust uniformly sampled in the range $[-T\_max, T\_max]$, where $T\_max$ denotes a rotor's max output thrust, to the drone's rotors for $1s$, before a new random thrust is applied. The soft drone would be dancing and twisting in the air, but despite that, this method leads to successful trainings, while generating data from a guiding controller, or switching random control signals at every instant, would fail. Secondly, we've found that trainings converge much better when we predict the next frame's velocity using the current one, instead of directly predicting the acceleration. Although this approach can lead to local minima for settling at the input, current velocity, that does not happen in our experiments and the acceleration can be convincingly reconstructed from the two velocities, as we shall elaborate in the results section. 

\textbf{Controlling the Learned Dynamics}\ \ \ \ \ 
Once the networks $\mathbf{d}$, $\mathbf{g}$ and $\mathbf{h}$ are trained, the dynamics can be expressed as 
\begin{equation}\label{eq:ode}
\mathbf{\dot{x} = f(x,u) = }
\begin{bmatrix} 
\mathbf{\dot{s}}\\
\mathbf{\ddot{s}}\\
\mathbf{\dot{e}}\\
\mathbf{\ddot{e}}\\
\mathbf{\dot{p}}\\
\mathbf{\ddot{p}}\\
\end{bmatrix}
=
\begin{bmatrix} 
\mathbf{\dot{s}}\\
\mathbf{\frac{1}{\alpha} (d(s,\dot{s},u)-\dot{s})}\\
\mathbf{\dot{e}}\\
\mathbf{\frac{1}{\alpha}(g(s,\dot{s},e,\dot{e},u)-\dot{e})}\\
\mathbf{\dot{p}}\\
\mathbf{\frac{1}{\alpha}(h(s,\dot{s},e,\dot{e},\dot{p},u)-\dot{p})}\\
\end{bmatrix},
\end{equation}
To control such learned dynamics, we summon the Linear Quadratic Regulator (LQR), a well-proven method for controlling rigid drones. Since LQR requires a linear system, we will perform first-order Taylor expansion around an operating point $\mathbf{(x^{*},u^{*})}$ with Jacobian matrices from automatic differentiation which can be done with PyTorch. In particular, we write:
\begin{equation}\label{eq:linearize}
\mathbf{\dot{x} = f(x,u)} \approx \mathbf{f(x^{*},u^{*})}+\mathbf{A(x-x^{*})+B(u-u^{*})},
\end{equation}
where $\mathbf{A} = \frac{\partial{\mathbf{f}}}{\partial{\mathbf{x}}}\big|_{\mathbf{x^*,u^*}}$, and $\mathbf{B} = \frac{\partial{\mathbf{f}}}{\partial{\mathbf{u}}}\big|_{\mathbf{x^*,u^*}}$, as shown in Equation \ref{eq:matrixAB}:

\begin{minipage}{.65\linewidth}
\begin{equation*}
\mathbf{A} = 
\begin{bmatrix} 
\mathbf{O} & \mathbf{I} & 
\mathbf{O} & \mathbf{O} & \mathbf{O} & \mathbf{O} \\
\frac{1}{\alpha}\mathbf{\frac{\partial {d}}{\partial s}} & \frac{1}{\alpha}\mathbf{(\frac{\partial d}{\partial \dot{s}}-I)} & 
\mathbf{O} & \mathbf{O} & \mathbf{O} & \mathbf{O} \\
\mathbf{O} & \mathbf{O} & 
\mathbf{O} & \mathbf{I} & \mathbf{O} & \mathbf{O} \\
\frac{1}{\alpha}\mathbf{\frac{\partial {g}}{\partial s}} & \frac{1}{\alpha}\mathbf{\frac{\partial {g}}{\partial \dot{s}}} & 
\frac{1}{\alpha}\mathbf{\frac{\partial {g}}{\partial e}} & \frac{1}{\alpha}\mathbf{(\frac{\partial {g}}{\partial \dot{e}}-I)} & \mathbf{O} & \mathbf{O} \\
\mathbf{O} & \mathbf{O} & 
\mathbf{O} & \mathbf{O} & \mathbf{O} & \mathbf{I} \\
\frac{1}{\alpha}\mathbf{\frac{\partial {h}}{\partial s}} & \frac{1}{\alpha}\mathbf{\frac{\partial {h}}{\partial \dot{s}}} & 
\frac{1}{\alpha}\mathbf{\frac{\partial {h}}{\partial e}} & \frac{1}{\alpha}\mathbf{\frac{\partial {h}}{\partial \dot{e}}} & \mathbf{O} & \frac{1}{\alpha}\mathbf{(\frac{\partial {h}}{\partial \dot{p}}-I)}
\end{bmatrix}_{\mathbf{x^{*}, u^{*}}}
\end{equation*}
\end{minipage}
\hfill
\begin{minipage}{.3\linewidth}
\begin{equation}
\label{eq:matrixAB}
    \mathbf{B} =\begin{bmatrix} 
    \mathbf{O}\\
    \frac{1}{\alpha}\mathbf{\frac{\partial \mathbf{d}}{\partial u}}\\
    \mathbf{O}\\
    \frac{1}{\alpha}\mathbf{\frac{\partial\mathbf{g}}{\partial u}}\\
    \mathbf{O}\\
    \frac{1}{\alpha}\mathbf{\frac{\partial \mathbf{h}}{\partial u}}\\
    \end{bmatrix}_{\mathbf{x^{*}, u^{*}}}.
\end{equation}
\end{minipage}

If we make the assumption that for $\mathbf{(x^{*},u^{*})}$ and $\mathbf{(x,u)}$ close enough to each other, $\mathbf{f(x,u) - f(x^{*},u^{*})} \approx \mathbf{ f(x-x^{*},u-u^{*})}$, then we have:
\begin{equation}\label{eq:proximity}
    \mathbf{\dot{(x-x^{*})} = f(x-x^{*},u-u^{*})} \approx \mathbf{A(x-x^{*})+B(u-u^{*})}.
\end{equation}

Once the linear system is obtained, LQR outputs the control matrix $\mathbf{K}$ and the control policy $\mathbf{u-u^{*} = -K(x-x^{*})}$ that drives $\mathbf{x}$ to $\mathbf{x^{*}}$ while keeping $\mathbf{u}$ close to $\mathbf{u^{*}}$ by minimizing the cost function $\int_{0}^{\infty}\mathbf{(x-x^{*})^TQ(x-x^{*})+(u-u^{*})^TR(u-u^{*})}dt$. The $\mathbf{Q}$ and $\mathbf{R}$ matrices are cost matrices used to manage the tradeoff between the two objectives. The optimization is done by solving the Continuous-time Algebraic Riccati Equation with SciPy.

\paragraph{Online Relinearization}
Traditionally, the operating point, which is the state-actuation pair $\mathbf{(x^{*},u^{*})}$ is chosen to be a fixed point such that $\mathbf{f(x^{*},u^{*})=0}$. For typical rigid drones, such fixed points can be obtained trivially, and yet, for deformable drones, without iteratively testing and optimizing with the neural dynamic system, it is generally not possible to know beforehand which set of rotor input would exactly balance the internal stress, viscous damping, and external gravity, or if such a balance exists. We see the fixed-point assumption as being overly strict and seek to circumvent it. The purpose of assuming $\mathbf{f(x^{*},u^{*})=0}$ is to turn the affine Equation. \ref{eq:linearize} into a linear one which LQR recognizes. In that case, $\mathbf{(x^{*},u^{*})}$ is time-invariant and indeed $
\mathbf{\dot{x} = (\dot{x-x^{*}}}) \approx \mathbf{0}+\mathbf{A(x-x^{*})+B(u-u^{*})}
$. Since in our case we cannot guarantee $\mathbf{f(x^{*},u^{*}) = 0}$, then $
\mathbf{(\dot{x-x^{*}}}) \neq \mathbf{A(x-x^{*})+B(u-u^{*})},
$ so directly running LQR with $\mathbf{A}$ and $\mathbf{B}$ would typically fail. However, by making the proximity assumption as in Equation. \ref{eq:proximity}, we can obtain the linear form locally. In other words, with $\mathbf{(x^{*},u^{*})}$ not being fixed points, we can still regulate close-enough neighbor states to it. 

\begin{minipage}{.5\linewidth}
\vspace{-1\baselineskip}
\begin{algorithm}[H]
	\caption{Online Relinearizing LQR} 
    \textbf{Input:} 
    $\mathbf{x}_{curr}$, $\mathbf{x}_{goal}$, $kp$, $kd$, $n$, $\mathbf{Q}$, $\mathbf{R}$
    \begin{algorithmic}[1]
        \State $\mathbf{u}_{curr}\gets \mathbf{0}$
        \State $iter\gets 0$
		\While {running}
		    \State update $\mathbf{x}_{curr}$, $\mathbf{\dot{x}}_{curr}$
		    \State $iter\gets iter + 1$
		    \If{$iter\bmod n = 0$}
    			\State $\mathbf{x}_{wp}\gets kp\cdot(\mathbf{x}_{goal}-\mathbf{x}_{curr})+kd\cdot \mathbf{\dot{x}}_{curr}$
    			\State $\mathbf{A},\mathbf{B}\gets  \mathcal{J}(\mathbf{d},\mathbf{g},\mathbf{h}, \mathbf{x}_{curr},\mathbf{u}_{curr})$   
    			\State $\mathbf{K} = LQR(\mathbf{A, B, Q, R})$
    			\State $\mathbf{u}_{oper} \gets \mathbf{u}_{curr}$
		    \EndIf
			\State $\mathbf{u}_{curr} \gets -\mathbf{K}(\mathbf{x}_{curr}-\mathbf{x}_{wp}) + \mathbf{u}_{oper}$
		\EndWhile
	\end{algorithmic}
\label{alg:alg}
\end{algorithm}
\end{minipage}
\hfill
\begin{minipage}{.45\linewidth}  
As in Algorithm. \ref{alg:alg}, our solution is built upon this observation. We initialize our drone with arbitrary $\mathbf{x}$, and $\mathbf{u=0}$. At each instant, we will linearize around the current $\mathbf{(x,u)}$. We run LQR with the linear system to obtain $\mathbf{K}$. In practice, we don't want to attract neighboring state to current state, but rather drive the current state to our goal state. Our strategy is that if we want to reach a state $\mathbf{x}_{wp}$ from $\mathbf{x}_{curr}$, then we will pretent to be at $(\mathbf{x}_{curr}-\mathbf{x}_{wp})$ trying to reach $\mathbf{x}_{curr}$, and compute $\mathbf{u} = -\mathbf{K}((\mathbf{x}_{curr}-\mathbf{x}_{wp})-\mathbf{x}_{curr}) = -\mathbf{K}(-\mathbf{x}_{wp})$. Given the current state $\mathbf{x}_{curr}$ and the goal state $\mathbf{x}_{goal}$, we calculate $\mathbf{x}_{wp}$ using a PD control: $\mathbf{x}_{wp}=kp\cdot(\mathbf{x}_{goal}-\mathbf{x}_{curr})+kd\cdot\mathbf{\dot{x}}_{curr}$. The control matrix will be used for $n$ timesteps before updated again.
\end{minipage}

\section{Experiments and Evaluation}
\label{sec:result}

\begin{figure}
    \centering
    \includegraphics[width=1.\textwidth]{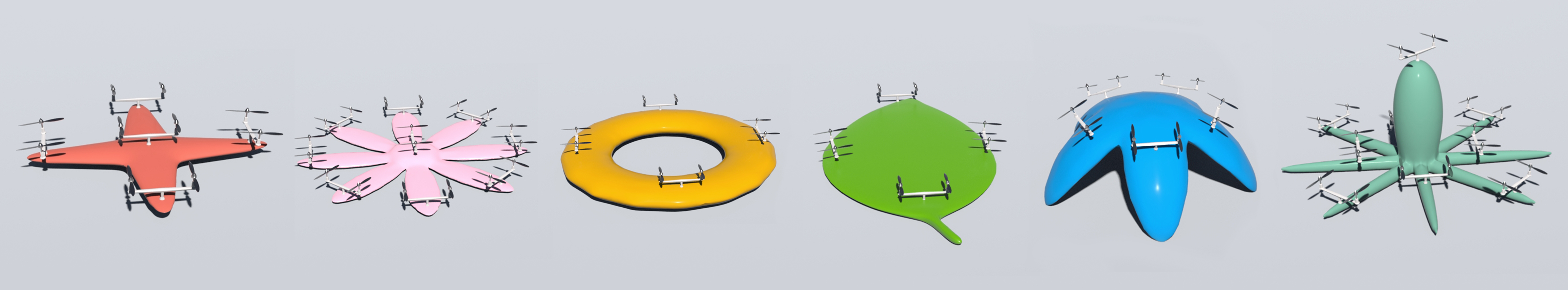}\hfill
    \caption{3D models used in our experiments}\label{fig:models}
\end{figure}

To test the efficacy of our method, we design a number of soft drone models in 2D and 3D featuring asymmetrical structures and odd numbers of rotors, as depicted in Figure. \ref{fig:models}. We conduct training on each model individually and use the generated controllers to direct the models to perform hovering, target reaching, velocity tracking, and active deformation.

\textbf{Testing the Linearized Networks}\ \ \ \ \ To verify the quality of our learned dynamics, we focus on two aspects. First, since our model predicts the next frame's velocity with the current velocity, rather than the acceleration (which is what we ultimately want), we need to make sure that the network actually learns how the velocity evolves, instead of reproducing the current velocity. As seen in the left subfigure of Figure. \ref{fig:networktesting}, which overlays the actual acceleration and the predicted acceleration reconstructed from the predicted velocity, the networks learn the evolution rules of velocity successfully. 
\begin{wrapfigure}[10]{r}{.6\textwidth}
 	\vspace{-1.4\baselineskip}
	\centering
    \includegraphics[width=0.3\textwidth]{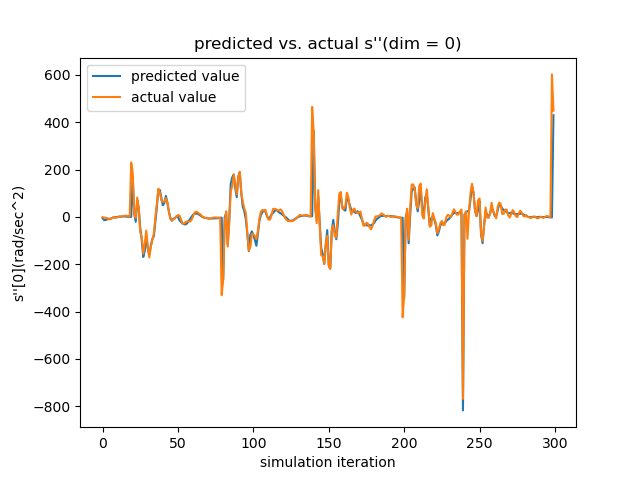}\hfill
    \includegraphics[width=0.3\textwidth]{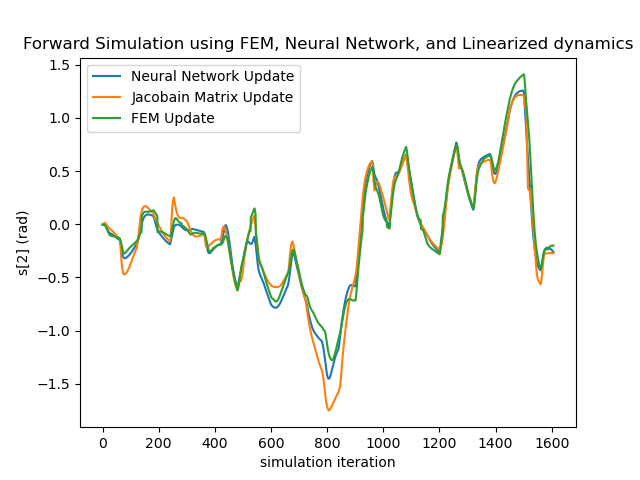}\hfill
    \\
    \caption{Testing results of the trained networks}\label{fig:networktesting}
\end{wrapfigure}
Secondly, since the LQR controller relies on the Taylor-expanded version of the learned networks, it is necessary to verify how well the networks' gradient matches that of the actual dynamics. Since the ground-truth gradient is difficult to obtain, we test this by reproducing temporal sequences using the linearized version as in Equation. \ref{eq:linearize} relinearized at $20Hz$. As shown in the right subfigure, even with error accumulation, the linearized evolution sequence (\textit{Jacobian Matrix Update}) keeps up with the ground truth for over a minute, showing the linearized neural network can reliably approximate the real dynamics locally.

\begin{figure}
    \includegraphics[width=0.4\textwidth]{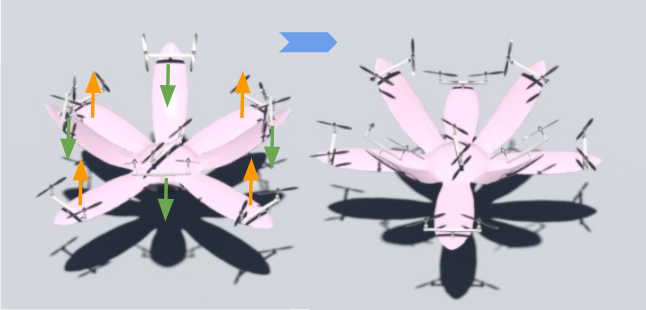}\hfill 
    \includegraphics[width=0.3\textwidth]{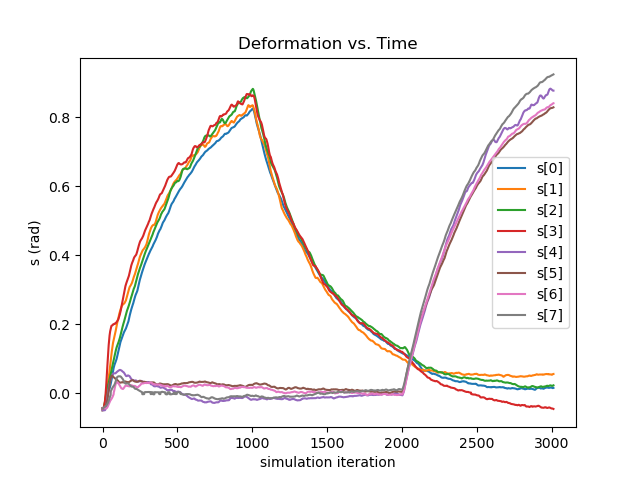}\hfill
    \includegraphics[width=0.3\textwidth]{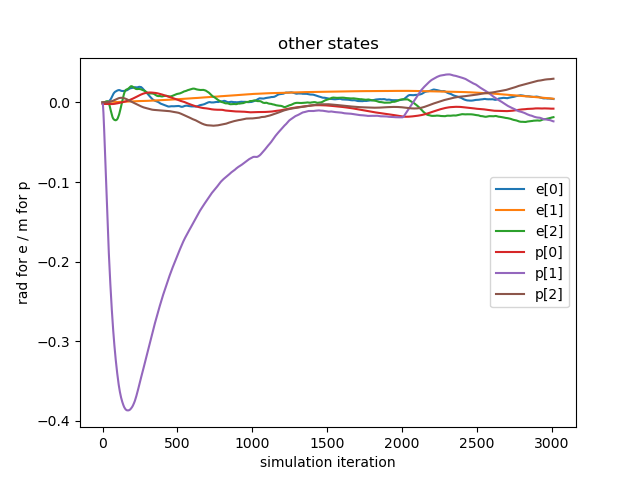}\hfill
    \caption{Active deformation while hovering}
\label{hover_deform}
\end{figure}

\textbf{Aerial Deformation}\ \ \ \ \ Our method successfully controls the drones to deform into specified configurations while hovering or tracking a velocity. The first experiment instructs the \textit{flower} drone to maintain its location and orientation while deforming into two different configurations: first with lateral petals raised and axial petals flat, then with lateral petals flat and axial petals raised, as shown in Figure. \ref{hover_deform}.  
In the middle subfigure, the pitch of all eight petals are controlled as expected in a smooth manner: petals $1,\ 3,\ 5,\ 7$ will rise first while petals $2\ ,4\ ,6\ ,8\ $ stay still, before the roles are gradually switched. The right subfigure shows that this is done while maintaining precise control of balance and position. Ever since the initial drop in the $-Y$ direction immediately after release, other rotational and translational movements are contained within $\pm 3^{\circ}$ and $\pm 5cm$. 
 
\begin{figure}
    \includegraphics[width=1\textwidth]{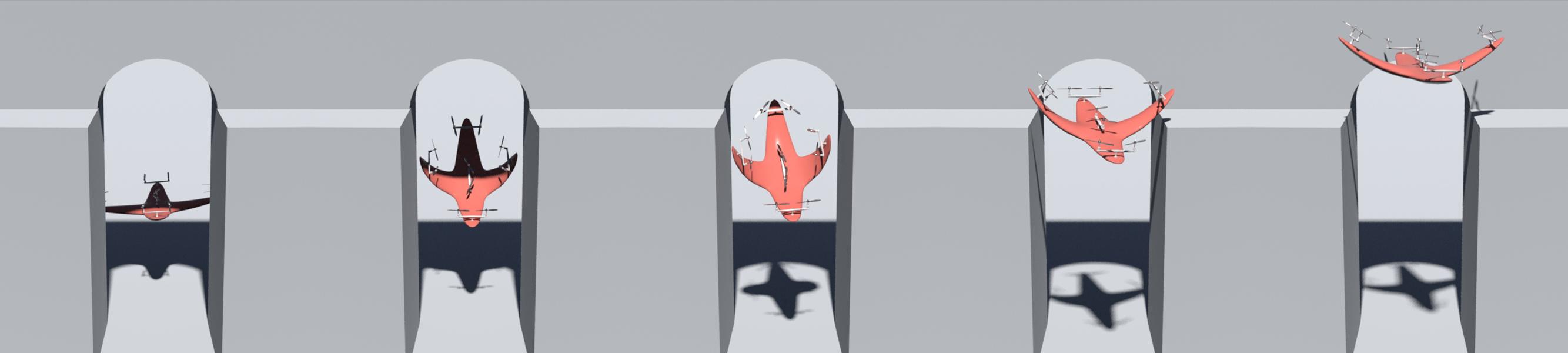}\hfill
     \includegraphics[width=1\textwidth]{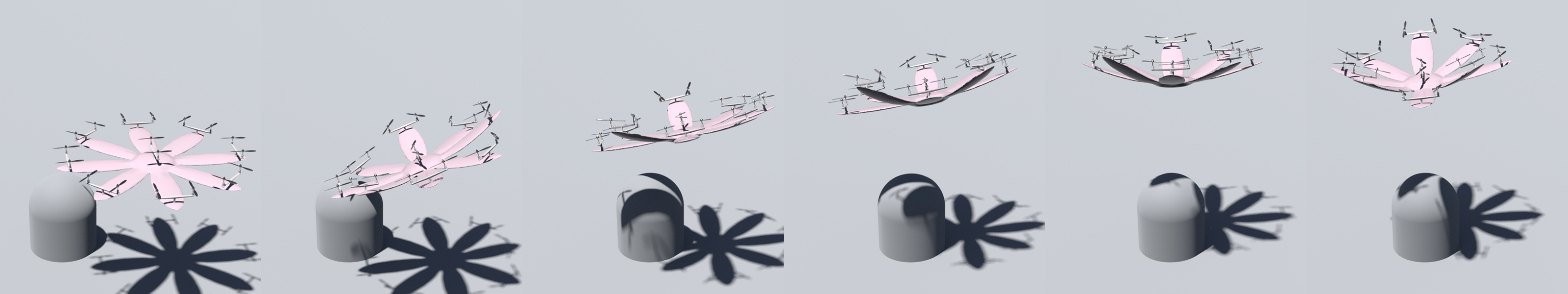}\hfill
    \caption{ Top: obstacle avoidance; bottom: Target reaching }\label{fig:animations}
\end{figure}

The second experiment combines velocity tracking with aerial deformation in an \textbf{obstacle avoidance} scenario. As shown in the top of Figure. \ref{fig:animations}, three concrete blocks form a gap that is narrower than the \textit{star} drone's body along the $Z$ direction. Here, we instruct the drone to fold up two wings by increasing their associated angles, and at the same time maintaining a forward and upward velocity in the $(+X, +Y)$ direction. It is also crucial to limit the deviation along the $Z$ direction in order to not crush into the blocks. As one sees in the first subfigure
\begin{wrapfigure}[10]{r}{.6\textwidth}
 	\vspace{-0.4\baselineskip}
	\centering
    \includegraphics[width=0.3\textwidth]{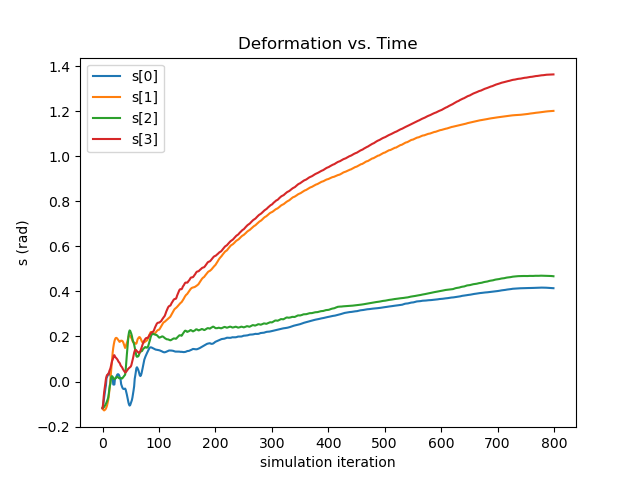}\hfill
    \includegraphics[width=0.3\textwidth]{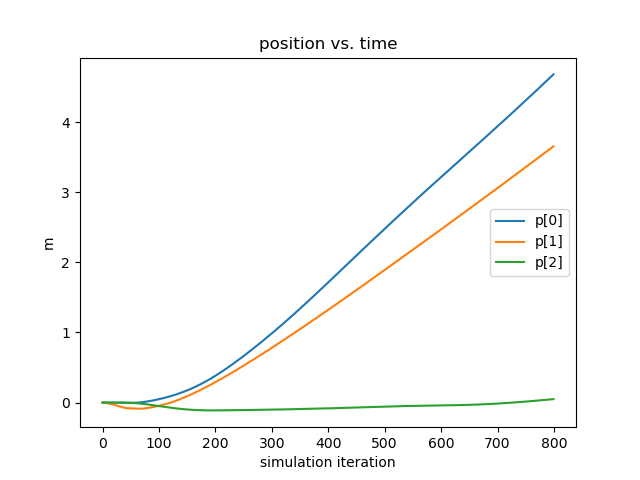}\hfill
    \\
    \caption{Obstacle avoidance performance}\label{fig:obstacle}
\end{wrapfigure}
 of Figure. \ref{fig:obstacle}, our controller increases the wing pitch by over $73^{\circ}$, reducing its width for over 30\%. As seen in the second subfigure, it does so while maintaining a steady velocity in the $(+X, +Y)$ direction, and deviating for less than $5cm$ in the $Z$-axis throughout the entire sequence, thus leading to the successful object avoidance as depicted in the bottom of Figure. \ref{fig:animations}.


\begin{figure}
    \includegraphics[width=0.25\textwidth]{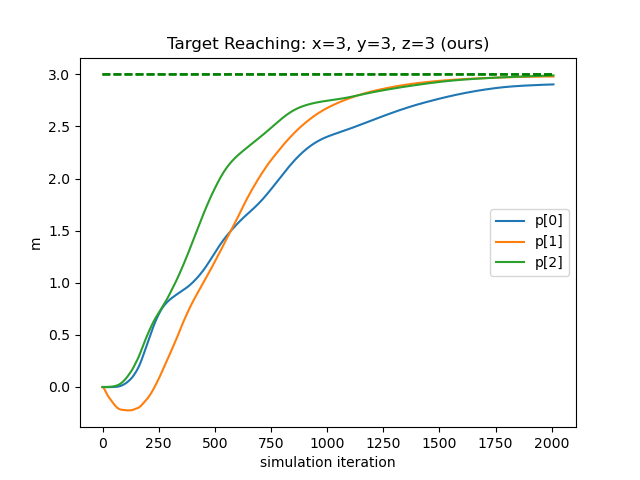}\hfill
    \includegraphics[width=0.25\textwidth]{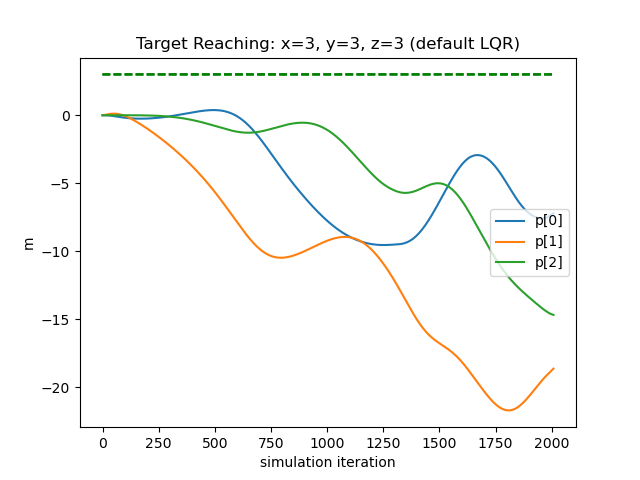}\hfill
    \includegraphics[width=0.25\textwidth]{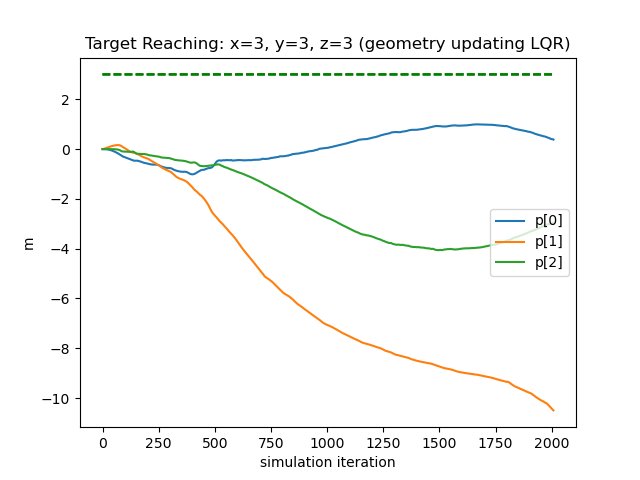}\hfill
    \includegraphics[width=0.25\textwidth]{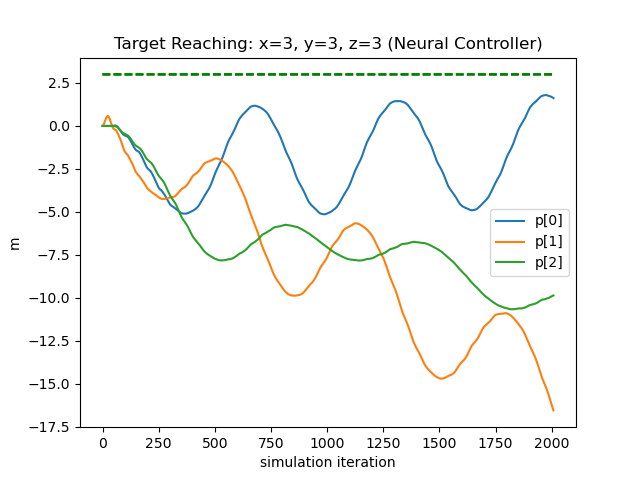}\hfill
    \caption{Left to right: ours, LQR, geometry-updating LQR, neural controller drive the drone to reach coordinate $(3,3,3)$ from $(0,0,0)$, plotted are the time-varying $(x,y,z)$ coordinates}\label{fig:sota}
\end{figure}

\textbf{SOTA Comparison}\ \ \ \ \ We compare our method with three different controllers: traditional LQR, geometry-updating LQR, and a neural network controller trained end-to-end using the strategy proposed by \citep{spielberg2019learning}. Our metrics will base on locomotion, since existing controllers for soft drone deformation is unavailable. The first candidate computes an LQR control matrix using the relevant quantities ------ fixed point, rotor position, rotor orientation, and rotational inertia, calculated from the drone's rest shape. The second recomputes the control matrix at every timestep with the relevant quantities updated based on the deformation, with further details given in the supplement. The third takes our trained network dynamic system as a differentiable simulator to train another network controller. We instruct all controllers to direct the \textit{flower} drone to travel from position $(0,0,0)$ to $(3,3,3)$. 
As depicted in Figure. \ref{fig:sota}, our method (left) successfully drives the $(x,y,z)$ coordinates of the drone to the target within 7\% error, while the other controllers fail due to the fragility of the soft drone dynamics. As shown in the table below, we compare their performance numerically with three metrics: final error, thrust usage, and the survival time before illegal configurations are encountered, which shows that our method excels the other candidates by far. 
\begin{center}
\begin{tabular}{ c|c|c|c|c } 
 \hline
 \multicolumn{5}{c}{Target Reaching} \\
 \hline
 metrics&ours&LQR&Geometry-updating LQR&Neural Controller\\
 \hline
 survival time (s) & \textbf{20.0} & 0.67 & 0.69 & 0.19 \\ 
 \hline
 final error (m) & \textbf{0.099} & 29.744 & 14.989 & 23.454\\
 \hline
 thrust usage (N) & \textbf{26740} & 90596 & 61947 & 179996\\
 \hline
\end{tabular}
\label{table: sota}
\end{center}

\begin{wrapfigure}[10]{r}{.3\textwidth}
 	\vspace{-1.2\baselineskip}
	\centering
 \includegraphics[width=0.3\textwidth]{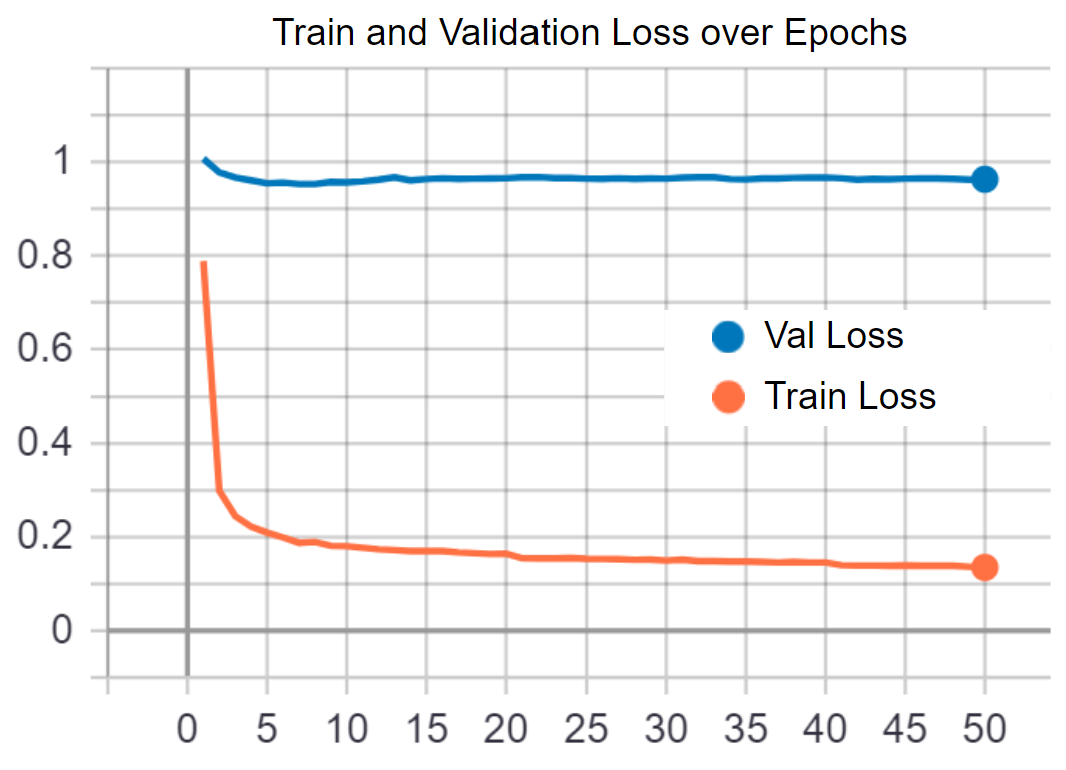}
    \\
    \caption{Convergence without decomposition}\label{fig:losseverything}
\end{wrapfigure}
\textbf{Ablation Testing: State Decomposition}\ \ \ \ \ The state decomposition is helpful for defining a local frame in which deformation can be expressed compactly as scalars. In this experiment, we train without decomposing and feed the network with the concatenated vector of the position, center orientation, and peripheral orientations. As displayed in Figure. \ref{fig:losseverything}, the validation loss is unable to converge successfully, signifying that the network fails to learn the correct patterns. This can be due to the fact that such unprocessed states double the size of the decomposed version, and also that the evolution rules of the position and orientation are highly unalike, thus demanding the network to evolve into multi-modal behaviors. 

\begin{figure}
    \includegraphics[width=0.25\textwidth]{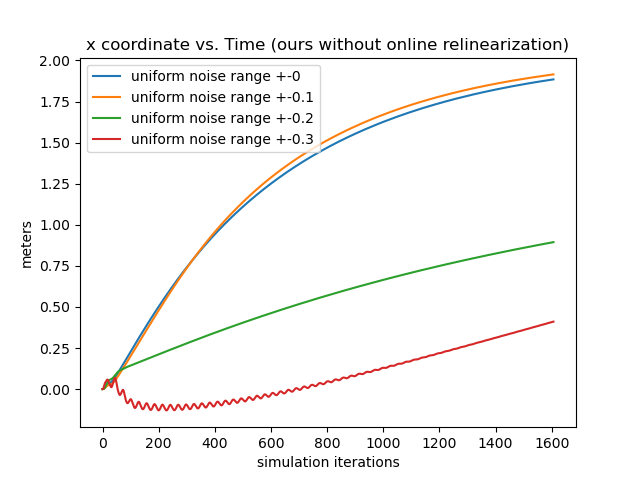}\hfill
    \includegraphics[width=0.25\textwidth]{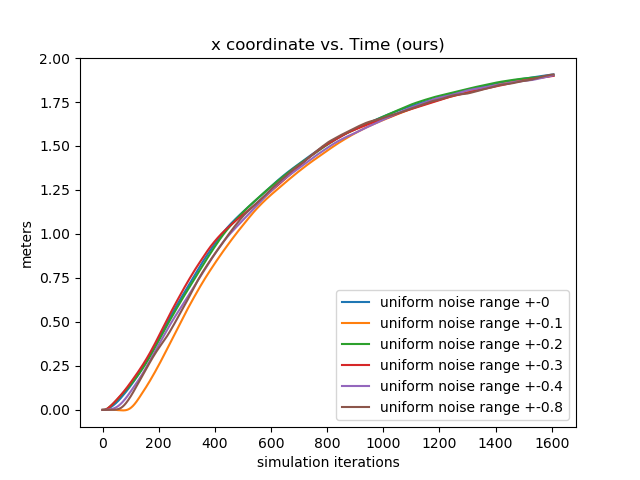}\hfill
    \includegraphics[width=0.25\textwidth]{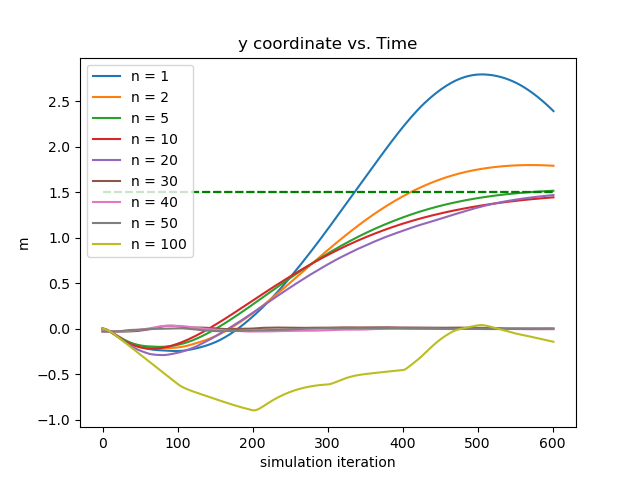}\hfill
    \includegraphics[width=0.25\textwidth]{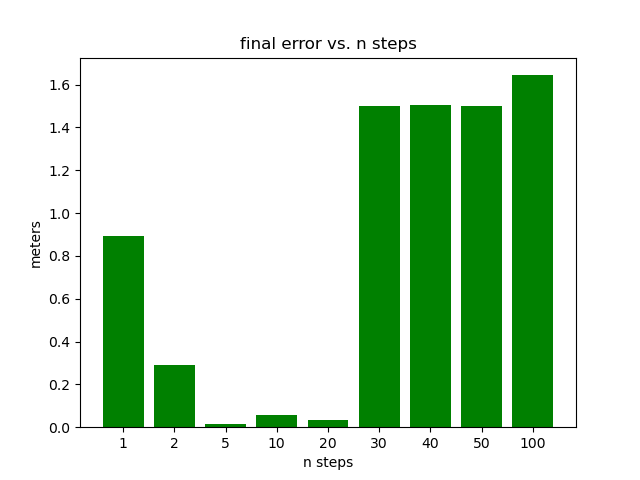}\hfill
    \caption{Left 2: The online relinearization scheme provides excelling robustness to non-fixed targets; Right 2: Control performance and final error under different relinearization timestep $n$}\label{fig:n}
\end{figure}
\textbf{Ablation Testing: Online Relinearization}\ \ \ \ \ 
We show the importance of the online relinearing mechanism by testing the controller performance where the input targets $\mathbf{(x}_{goal},\mathbf{u}_{goal})$ deviates from a fixed point $\mathbf{(x^*, u^*)}$ by different margins. We first obtain a fixed point by trial-and-error and then fine-tune it with gradient descent using our learned networks. Then, we keep the goal location untouched, and pollute the rest of the fixed point by adding uniform random noise proportional to each state variable by 10\%, 20\%, 30\%, 40\%, and 80\%. We direct the \textit{rod} drone to translate for 2 units along the $+X$ direction. The results are depicted in the left 2 subfigures of Figure. \ref{fig:n}. The first subfigure shows the results without online relinearization, where the control quality deteriorates significantly when over 20\% noise is added, and start to generate NANs after 40\% noise is added. The second subfigure shows the results with online relinearization, in which the controller performs steadily even with 80\% noise added. The significance is that during deployment the pilot only needs to input the target configuration without worrying about the hard-to-compute fixedness of such configuration, thereby making our control system feasible for human piloting.


\textbf{Effect of Relinearization Frequency}\ \ \ \ \ 
The parameter $n$ controls the relinearization frequency $\frac{100}{n}Hz$. The more often the system is linearized the more accurate the linear approximation is. And also, since we set a new waypoint at each linearization, whose distance away is inversely proportional to $n$, frequent linearzation sets many adjacent waypoints while infrequent linearization sets long-term, sparse waypoints. In this experiment, we test the performance of 9 different values of $n$: $\{1, 2, 5, 10, 20, 30, 40, 50, 100 \}$, and show that it is not the case that larger $n$ implies better performance. As shown in the right two subfigures of Figure. \ref{fig:n}, $[5,\ 20]$ is clearly the sweet-spot of this parameter, where smaller $n$ leads to overshoots, and larger $n$ insufficient actuation.



\section{Discussion and Conclusion}
\label{sec:conclusion}
We propose a computational system to generate controllers for soft multicopters that jointly controls the locomotion and active deformation, without relying on extra mechanical parts.
Our method takes advantage of a physics-inspired decomposed state space, and train neural networks to represent the dynamics. We control the neural dynamics system using an LQR controller enhanced with a novel online relinearization scheme. We use our method to successfully generate controllers for a variety of soft multicopters to perform hovering, target reaching, velocity tracking, and active deformation.

\textbf{Sim2Real Transfer} \ \ \ \ \ 
Our method is well-suited for real-world adaptations. First, we limit the interfacing between the simulator and the learning system strictly to sensor readings, so we make sure that no unrealistic benefit is gained from experimenting virtually. Secondly, we form a straightforward guideline for sensor deployment, featuring accessible gadgets with easy installation. Thirdly, the computation of LQR optimization and neural network evaluation can be realistically carried out in real-time by onboard computers, as previously explored by \citep{kaufmann2018deep, foehn2018onboardLQR}. Finally, being data-driven, our method betters the analytic approaches for Sim2Real adaptation, since it learns from data which contain the unmodelled nuances of the real world. The main challenge for Sim2Real transfer would be the experimental designs to generate meaningful data using the fabricated drones.

\textbf{Limitations}\ \ \ \ \ With our approach there are several limitations. First, the sensor placement requires human design, and there lacks a mechanism to tell, before training, if a sensing scheme would work. 
Secondly, every drone requires a separate training with no knowledge transfer. 
Thirdly, we use dual rotors with counter-rotation to cancel the spinning torque effects, which complicates the manufacturing process. 
Finally, the current control scheme is ineffective for aggressive maneuvers.

In the future, we plan to build an automated system for optimizing sensor locations as well as a unified neural dynamics platform that facilitates knowledge transfer, and to evaluate our approach on real-world soft multicopters.


\clearpage
\acknowledgments{We thank all the reviewers for their valuable comments. We acknowledges the funding support from Neukom Institute CompX Faculty Grant and Neukom Scholar Program, Burke Research Initiation Award, Toyota TEMA North America Inc, and NSF 1919647. We credit the Houdini Education licenses for the video generations.}


\bibliography{example}  
\clearpage


\vbox{%
\hsize\textwidth
\linewidth\hsize
\vskip 0.1in
\centering
{\LARGE\bf Soft Multicopter Control using\\ 
Neural Dynamics Identification \\
Supplementary \par}
\vskip 0.3in
}


\renewcommand\thesection{\Alph{section}} 
\setcounter{section}{0}

\section{Overview}
In this document, we present the supplementary materials to our published paper. In Section B, we describe the specifications of the 2D and 3D drone models used in our training and testing, including the soft material properties, the rotor designs, and the sensor placements. In Section C we introduce the details of how the state vector $\mathbf{s}$ and $\mathbf{e}$ are obtained from sensor readings. In Section D we propose a general guideline for deploying IMU sensors for arbitrary drone shapes. In Section E we describe the simulation environment, the simulation model used, and the noise treatment. In Section F, we specify the details of the learning module, including the network structure used, the data generation scheme, as well as the techniques and hyperparameters used along the training procedure. In Section G, we specify the parameters used in our control module and the mathematical derivation of the benchmark LQR controller. In Section H, we present a discussion about the system design choices, the assumptions we have made, and the potential challenges for the fabrication and control of real-world soft multicopters. 

\section {Drone Designs}
\begin{figure}[hbt!]
    \centering
    \includegraphics[width=0.33\textwidth]{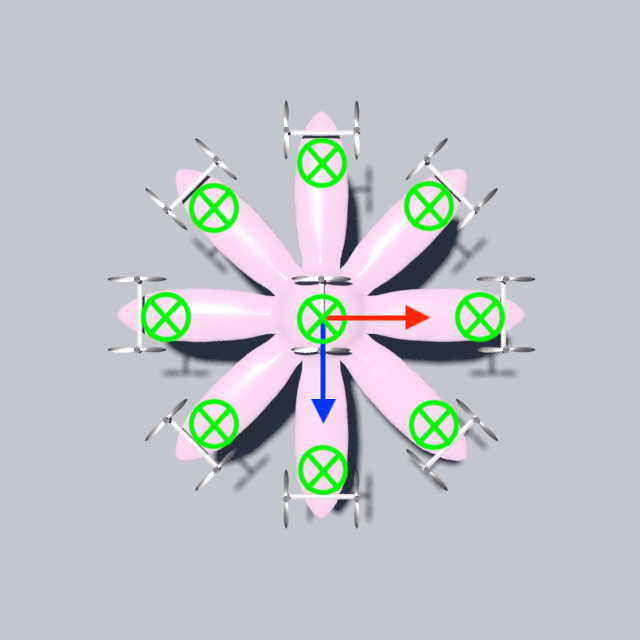}\hfill
    \includegraphics[width=0.33\textwidth]{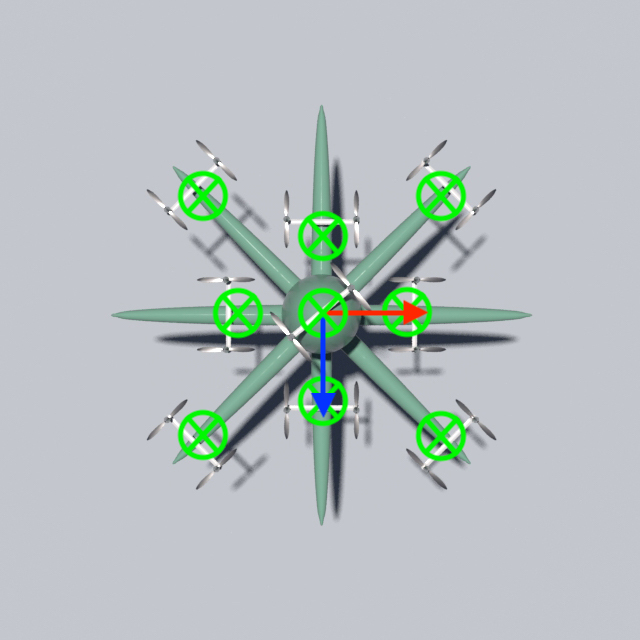}\hfill
    \includegraphics[width=0.33\textwidth]{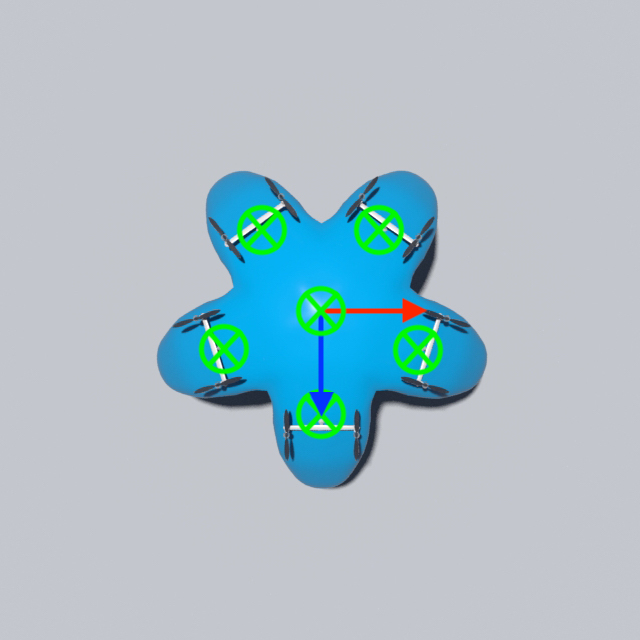}\hfill
    \includegraphics[width=0.33\textwidth]{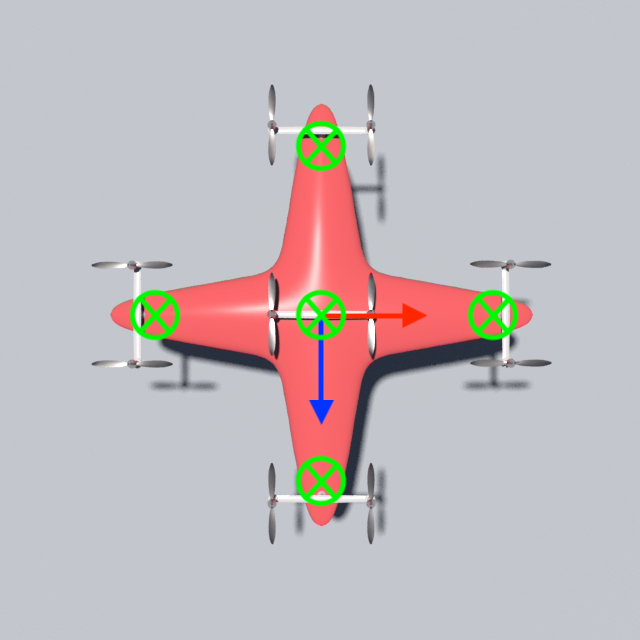}\hfill
    \includegraphics[width=0.33\textwidth]{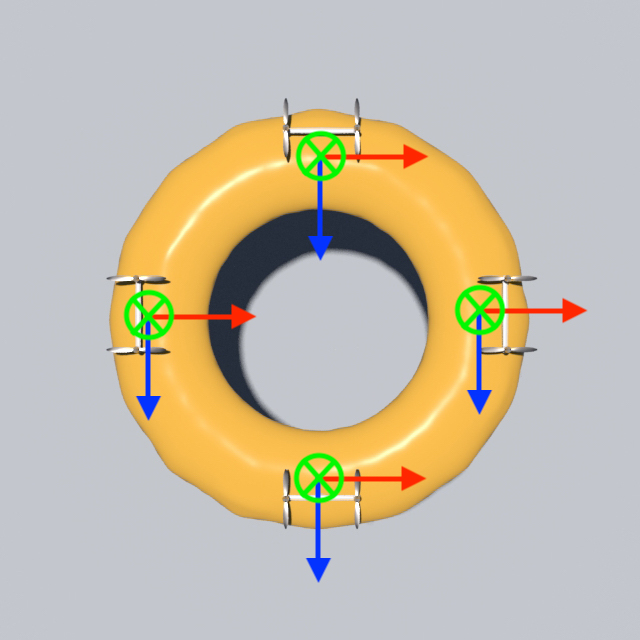}\hfill
    \includegraphics[width=0.33\textwidth]{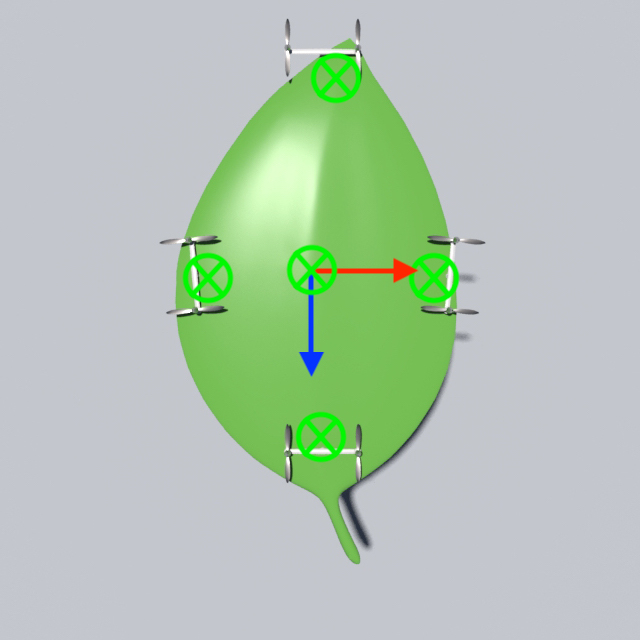}\hfill
    \caption{Sensor placement for 3D drone designs. Top row: Flower, Octopus, Orange Peel; Bottom row: Star, Donut, Leaf.}\label{fig:3ddroneimu}
\end{figure}

\begin{figure}[hbt!]
    \centering
    \includegraphics[width=0.2\textwidth]{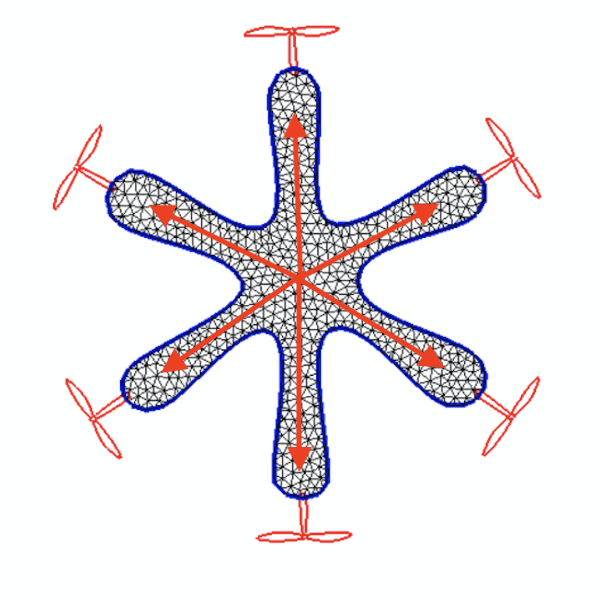}\hfill
    \includegraphics[width=0.2\textwidth]{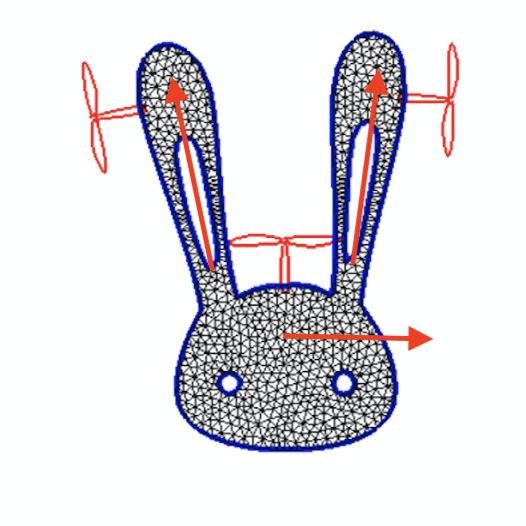}\hfill
    \includegraphics[width=0.2\textwidth]{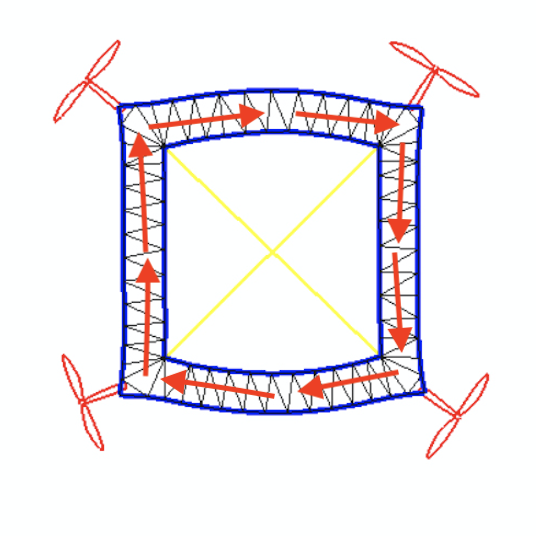}\hfill
    \includegraphics[width=0.2\textwidth]{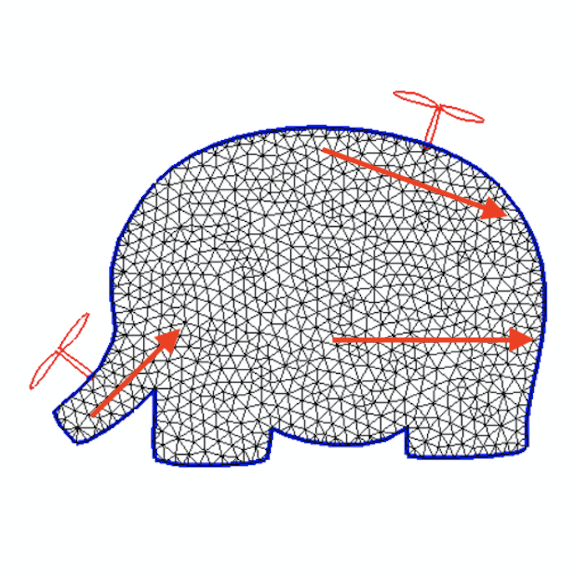}\hfill
    \includegraphics[width=0.2\textwidth]{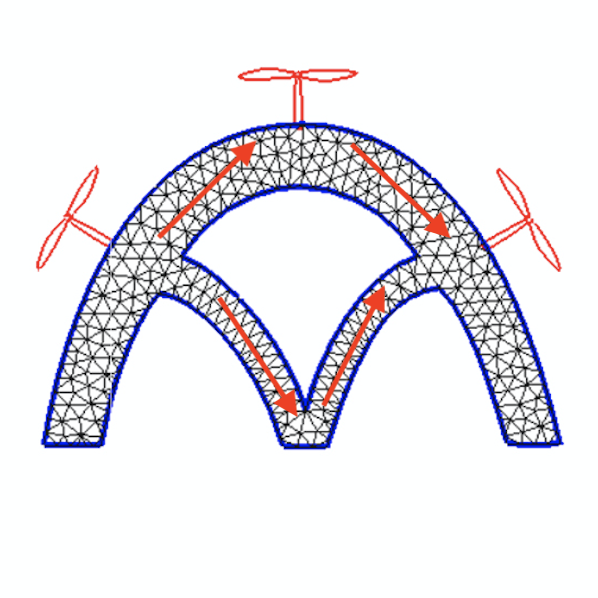}\hfill
    \includegraphics[width=0.7\textwidth]{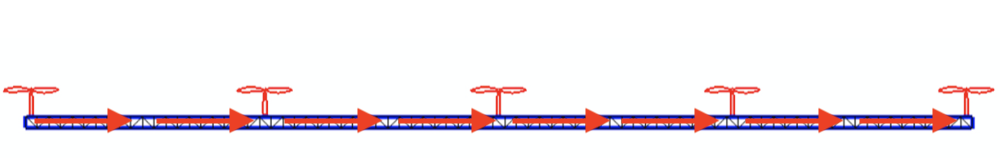}\hfill
    \caption{Sensor placement for 2D drone designs (illustrated by red arrows). Top row: Engine, Bunny, Diamond, Elephant, Rainbow; Bottom row: Rod.}\label{fig:2ddroneimu}
\end{figure}

\paragraph{Sensor Layouts} 
For the 3D examples, the sensing scheme is depicted in Figure. \ref{fig:3ddroneimu}. Each IMU measures the local $X$, $Y$, and $Z$ axes, which are coded by Red, Green, and Blue respectively. The $Y$ axis will point out of the plane. For the peripheral measurements, we will only make use of the measured $Y$ axis.
Since rotation in 2D can be represented by one scalar only, for 2D drones the IMU will only output the angle between the measured vector and the horizontal. The measured vectors are depicted in Figure. \ref{fig:2ddroneimu}. 
Please also refer to Figure. \ref{fig:3ddroneimu} and Figure. \ref{fig:2ddroneimu} for the nicknames of these drone models. For 2D drones, we only train controllers of the \textit{rod} model since it is the most deformable 2D geometry among all and therefore the one that displays the most interesting behaviors.

\paragraph{Drone Specifications} The specifications of our tested models' size and material properties are presented in Table. \ref{tab:forms}.


\begin{figure}[t]
    \centering
    \includegraphics[width=0.48\textwidth]{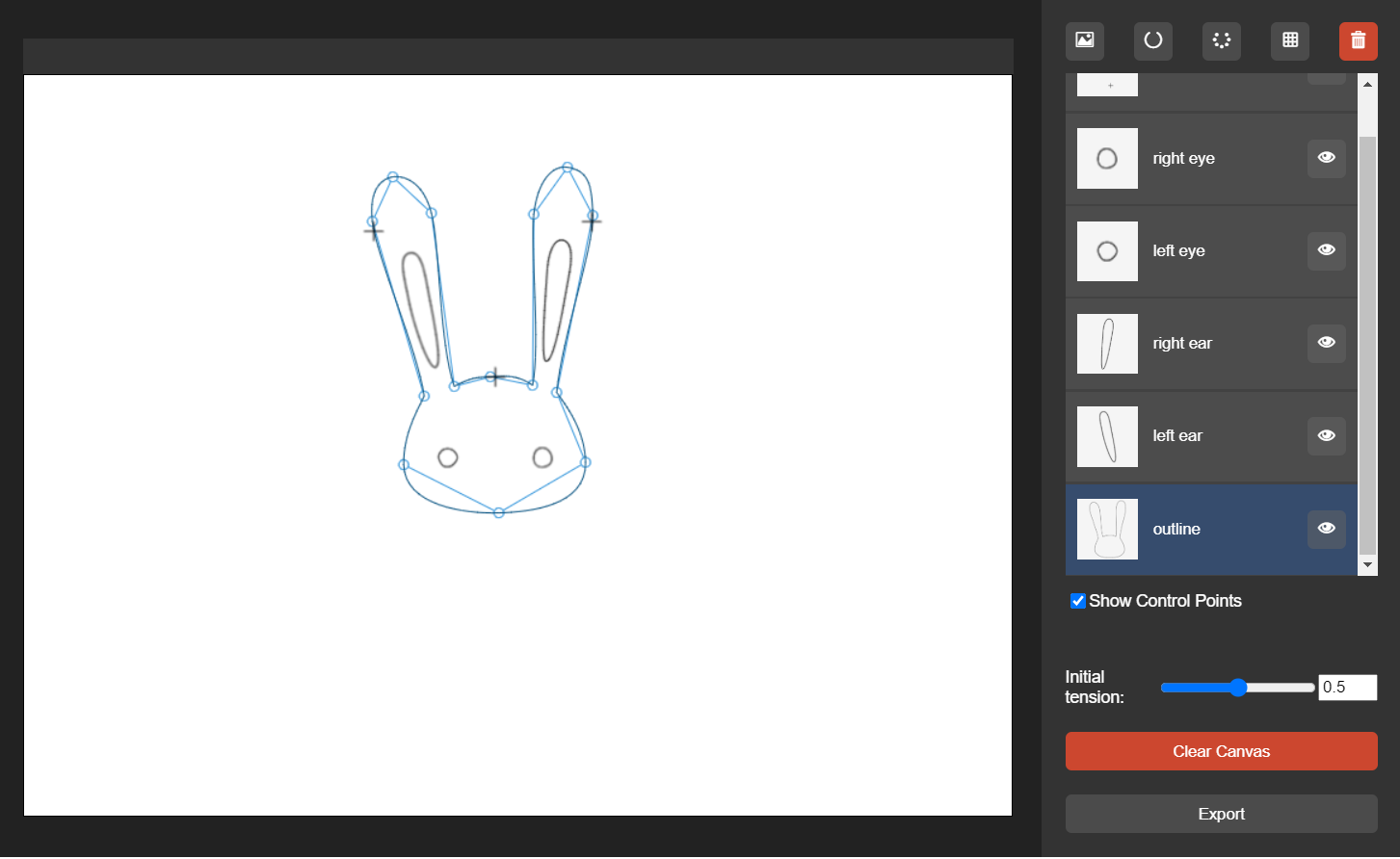}\hfill
    \includegraphics[width=0.48\textwidth]{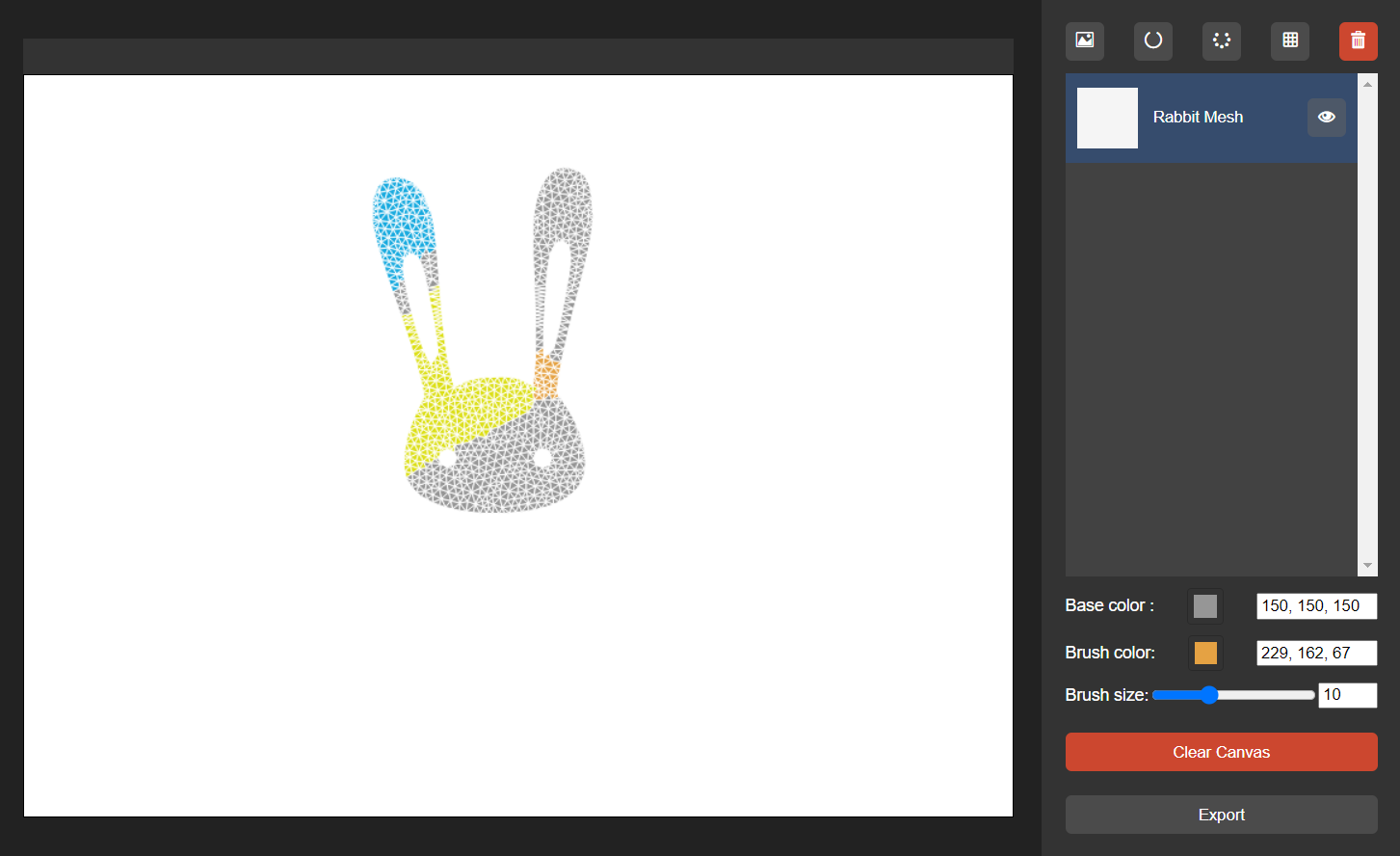}\hfill
    \caption{Left: sketching the contour; Right: painting the mesh.}\label{fig:ui}
\end{figure}

\paragraph{Drone Design Procedure} To customize 2D drones, we develop a web-based painting tool , as shown in Figure. \ref{fig:ui}, to sketch the contours, and use TetGen\citeS{tetgen} to create triangle meshes from the contours. The interface also allows users to set rotor positions and assign materials to triangle elements of the mesh interactively. 3D drones are modeled in Maya and then converted to tetrahedron meshes using TetGen.

\paragraph{Dual-Propeller Rotor}
A rotor mounted on a soft drone will influence the drones' body with
\begin{enumerate}
    \item the thrust from accelerating the air and creating a low-pressure region in front of it, a force which will act in the normal direction of the surface on which the rotor is mounted;
    \item the torque that acts on the drone's body in the opposite direction of the rotor's rotation to conserve angular momentum;
    \item the gyroscopic torque that will act in the direction perpendicular to the gravity and the rotor's spinning direction, which happens when the rotor is tilted.
\end{enumerate}
In this work, each of the $m$ actuators will be implemented by a dual-rotor with counter-rotation, and the actuation will be split in half for each of the two rotors. With the two sub-rotors spinning in countering directions, the second term will be canceled out. The two sub-rotors will cancel the gyroscopic moments of their counterparts as well. Under this setting, in our simulation, only the normal force is modeled.

\begin{table}
\centering
\begin{tabular}{ ccccccc } 
 \hline
 \multicolumn{7}{c}{3D models} \\
 \hline
 specs&Donut&Starfish&Flower&Leaf&Octopus&Orange peel\\
 \hline
 mass($kg$) & 1 & 1 & 1 & 1 & 1 & 1\\ 
 \hline
 modulus($N/m^{2}$) & 1e4 & 3e3 & 6e3 & 3e3 & 1e4 & 5e2\\
 \hline
 length-x($m$) & 3.5 & 3.6 & 3.6 & 2.4 & 3.6 & 3\\
 \hline
 length-y($m$) & 0.36 & 0.375 & 0.225 & 0.075 & 1.5 & 1.3\\
 \hline
 length-z($m$) & 3.5 & 3.6 & 3.6 & 4.5 & 3.6 & 2.9\\
 \hline
 num sensors & 4 & 4 & 8 & 4 & 8 & 5 \\
 \hline
 num rotors & 4 & 5 & 9 & 4 & 9 & 5\\
 \hline
 max thrust($N$) & 10 & 10 & 10 & 10 & 10 & 10\\
 \hline
\end{tabular}

\begin{tabular}{ccccccc} 
 \hline
 \multicolumn{7}{c}{2D models} \\
 \hline
 specs&Engine&Bunny&Diamond&Elephant&Rainbow&Long Rod\\
 \hline
 mass($kg$) & 1 & 1 & 1 & 1 & 1 & 1 \\ 
 \hline
 modulus($N/m^{2}$) & 6e3 & 6e3 & 6e3 & 6e3 & 6e3 & 6e3 \\
 \hline
 length-x($m$) & 1.90 & 1.12 & 1.47& 2.46 & 2.08 & 0.1 \\
 \hline
 length-y($m$) & 2.16 & 1.75 & 1.42& 1.69 & 1.30 & 8.0 \\
 \hline
 num sensors($m$) & 6 & 3 & 8 & 3 & 4 & 8 \\
 \hline
 num rotors($m$) & 6 & 3 & 4 & 2 & 2 & 5 \\
 \hline
 max thrust($N$)& 10 & 10 & 10 & 10 & 10 & 10 \\
 \hline
\end{tabular}
\caption{\label{tab:forms}Design Specifications}
\end{table}

\section{Computation of the State Vectors}

\begin{figure}[t]
    \centering
    \includegraphics[width=0.5\textwidth]{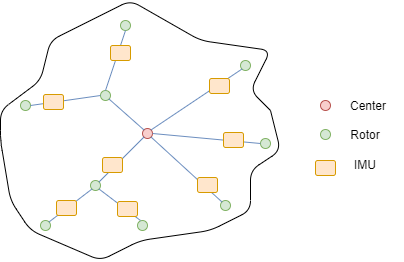}\hfill
    \includegraphics[width=0.4\textwidth]{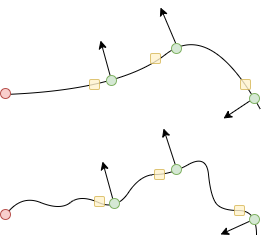}\hfill
    \caption{IMU placement. The left figure describes the proposed scheme for inserting IMUs for drones with arbitrary, irregular geometries. The right illustrates the different level of adequacy of this scheme at two different levels of softness.}\label{fig:imu}
\end{figure}

\paragraph{Computation of e}
In the common case where the drone's body contains no hole in the middle, an IMU will be placed at the geometric center, and the measured rotation of the IMU's rigid frame will be used as the definition of the drone's body frame. For cases like the \textit{donut}, where there is a hole in the middle, the strategy is to insert a few IMU at the circumferential locations, and average these obtained rotations. In our case where the 4 inserted IMUs are center-symmetric, the average rotation is obtained by averaging the body-frame $X$-direction of IMU 1, 3, the body-frame $Y$-direction of IMU 2, 4, and use cross products to obtain the combined body frame. For the general case, this operation can be done by converting these measurements into quaternions and apply the averaging methods described in ~\citeS{2markley2007averaging} to
obtain the body frame.

\paragraph{Computation of s}
The deformation vector $\mathbf{s}$ will constitute measurements from IMUs inserted at peripheral points. For measuring these local deformations, we will measure the normal vector of the local body surface, which is the direction of the $Y$-axis of IMU's body frame. 
Given the IMU's measured rotation matrix (body-to-world) $\mathbf{R}_{peripheral}$, we will first calculate its $Y$-axis in the world frame by 
\begin{equation}
\mathbf{y}_w = \mathbf{R}_{peripheral}
\begin{bmatrix}
0\\
1\\
0\\
\end{bmatrix}.
\end{equation}
Then given the body-to-world rotation matrix $\mathbf{R}_{central}$ defined by $\mathbf{e}$, we will map the $\mathbf{y}_w$ on to the drone's body frame:
\begin{equation}
\mathbf{y}_b = \mathbf{R}_{central}^T\mathbf{y}_w
\end{equation}
Then, an axis-angle will be calculated for how to rotate the Y-axis in the drone's body frame to $\mathbf{y}_b$. 
The axis will be calculated by:
\begin{equation}
\mathbf{v} = \mathbf{y}_b\times\begin{bmatrix}
0\\
1\\
0\\
\end{bmatrix}
\end{equation}
\begin{equation}
\mathbf{\hat{v}} = \frac{\mathbf{v}}{||\mathbf{v}||}
\end{equation}
The angle will be calculated by:
\begin{equation}
\alpha = \beta \cdot arccos(\frac{\mathbf{y}_b}{||\mathbf{y}_b||}\cdot\begin{bmatrix}
0\\
1\\
0\\
\end{bmatrix})
\end{equation}
where 
\begin{equation}
  \text{$\beta$} =
  \begin{cases}
                                  1 & \text{if $\hat{v}\times\mathbf{r}$ has positive x-entry} \\
                                  -1 & \text{if $\hat{v}\times\mathbf{r}$ has negative x-entry} \\
  \end{cases},
\end{equation}
with $\mathbf{r}$ representing the body frame location of the inserted IMU when undeformed.

In this way, the deformation is converted into a scalar, and by the construction of $\beta$, the magnitude of the scalar will represent the magnitude of the deformation, while the sign represents whether the deformation is inward (positive) or outward (negative).

\section{Guidelines for Sensor Placement}
We present a general guideline for selecting where IMUs are deployed in the left part of Figure. \ref{fig:imu}. Given an arbitrary drone shape in 3D (pressed onto the X-Z plane), we build a tree with the root node being the geometric center of the drone, and the child nodes being the rotors. The IMUs will be inserted at the edges of the tree near the outer rotor. The effectiveness of this approach is contingent on the simple modality of the soft drone's deformation. For instance, if you take a look at the right part of Figure. \ref{fig:imu}, for the above case, the deformed shape of the drone's arm can be approximately reconstructed from the three measurements, whereas in the lower case, the three measurements are far from enough to describe the deformed shape, as the deformation is highly multi-modal, while these higher-order deformation modes are effectively beyond the controlling capacity of the drone's rotors. 
As a result, it is the task in the design of these drones (mostly selecting the modulus and thickness) so that the drone is soft enough to perform significant deformation, while the deformation mode of the drone is simple. In practice, this IMU insertion guideline works for our various examples.

\section{Simulation Description}
\paragraph{Soft Body Model}
In our simulation environment, the deformation of a soft body is simulated using an explicit co-rotated elastic finite element model \citeS{2muller2004interactive}. A mass-proportional damping term is used to model the damped elastic behavior. We use tetrahedron (3D) and triangle (2D) meshes for discretization. An OpenMP-based parallel implementation of the elastic solver was employed to boost the simulation performance. Each rotor is rigidly bound to a local set of surface vertices on the finite element mesh in the course of the simulation, with the rotor direction aligned to the averaged normal direction of the local surface triangle primitives in 3D (surface segments in 2D).  

In our simulation environment, an IMU is implemented by binding a number of nearby vertices and use their positions to define a reference frame via cross products.

\paragraph{Noise Treatment}
In the simulation environment, in order to emulate the perturbations and uncertainties in the real world, noise is added to the sensor readings, and a time delay is added to the rotor output. The details of these noises are given in the table below.
\begin{center}
\begin{tabular}{ lll } 
 \hline
 Category & Noise type & Level \\ 
 \hline
 angle measurements & Gaussian & $\mu = 0$, $\sigma = 0.573^\circ$\\
  \hline
 position measurements & Gaussian & $\mu = 0$, $\sigma = 0.01 m$\\
  \hline
 rotor perturbation & Gaussian & $\mu = 0$, $\sigma = 0.1 N$ \\
 \hline
 output delay & constant & $0.03 s$ \\
 \hline
\end{tabular}
\end{center}


\section{Learning of Neural Networks}

\paragraph{Dataset Generation}
The training data are generated with our implementation of a Finite Element simulator. Given a drone geometry, we initialize the drone as undeformed, lying at the origin, and apply a random thrust to each rotor and observe the drone's position, rotation, and deformation at $100Hz$. Each set of random thrust is applied for $1s$. Other data generation schemes we tried also consist of using a rigid LQR controller to generate the thrusts, or apply a different random thrust each frame, but the former yields poor test loss due to the confined distribution of LQR control outputs, while the latter generates data too noisy to train on. The insight is that we need to give the system enough time to respond to a signal and display meaningful behaviors.

\begin{figure}
    \centering
    \includegraphics[]{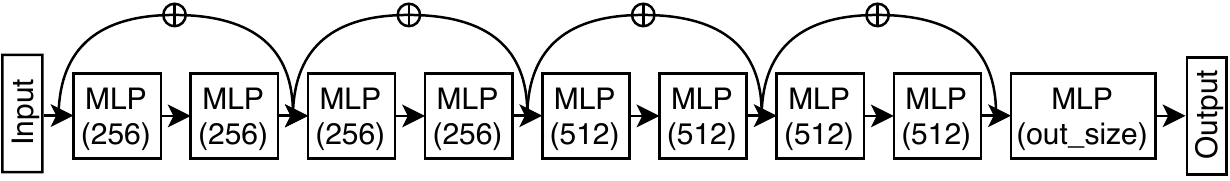}\hfill
    \caption{Network architecture: This architecture is similar to ResNet. Here each MLP consists of a fully connected layer and a ReLU except the last layer. The number in the parentheses means output dimension of each layer.}\label{fig:net}
\end{figure}

\paragraph{Network Architecture and Training}
As shown in Figure. \ref{fig:net}, All the three neural networks to learn $\{\mathbf{d}, \mathbf{g}, \mathbf{h}\}$ adopts the same architecture. The architecture is similar to ResNet except that the convolution layers are replaced by fully connected layers. Note that there are no normalization techniques used in our networks.
We use Adam optimizer with initial learning rate 0.001 and decay rate 0.8 for each 20 steps. The batch size is 512. We train for 50 epoches. For loss function we found out L1 loss provides superior result to L2 loss due to the robustness of the L1 loss.

\begin{figure}
    \centering
    \includegraphics[width=0.33\textwidth]{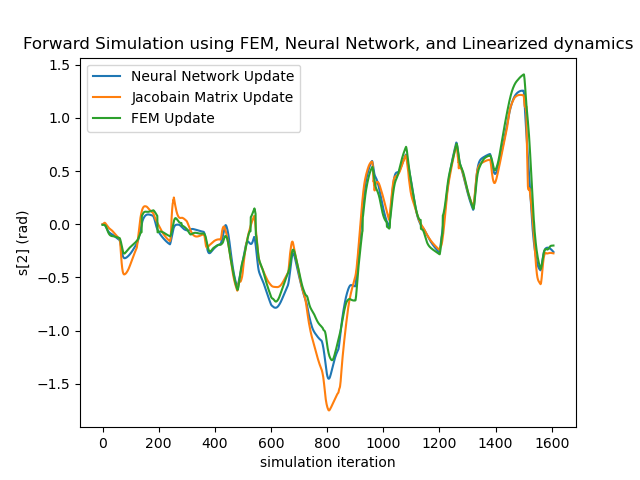}\hfill
    \includegraphics[width=0.33\textwidth]{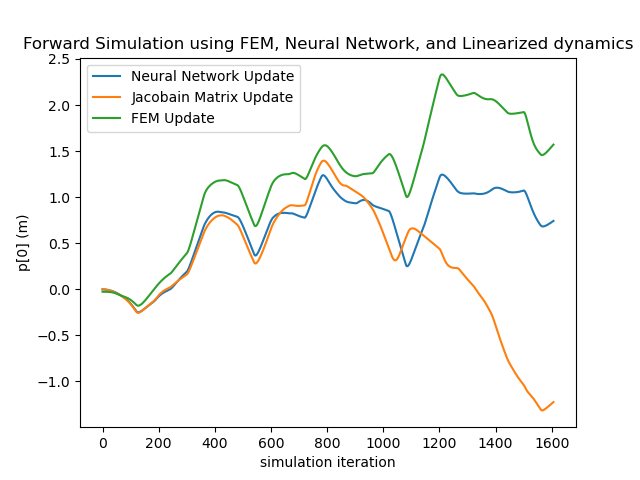}\hfill
    \includegraphics[width=0.33\textwidth]{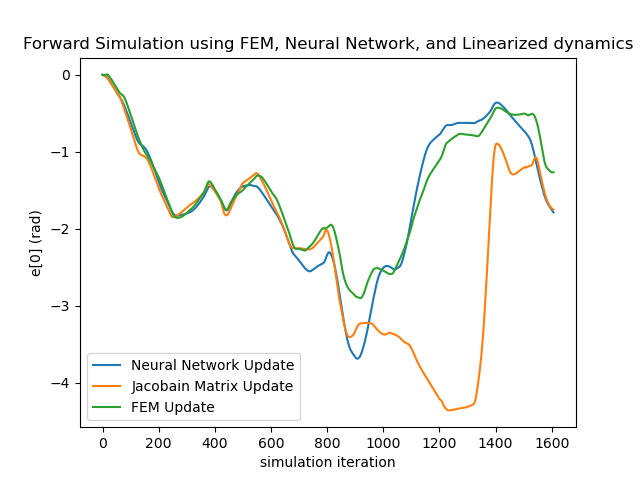}\hfill
    \caption{Testing results of the Network Linearization}\label{fig:jacobtesting}
\end{figure}

\paragraph{Testing of the Networks}
\begin{figure}
    \centering
    \includegraphics[width=0.33\textwidth]{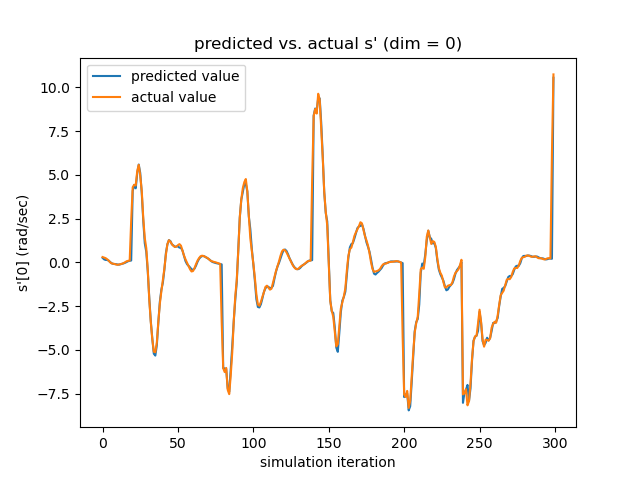}\hfill
    \includegraphics[width=0.33\textwidth]{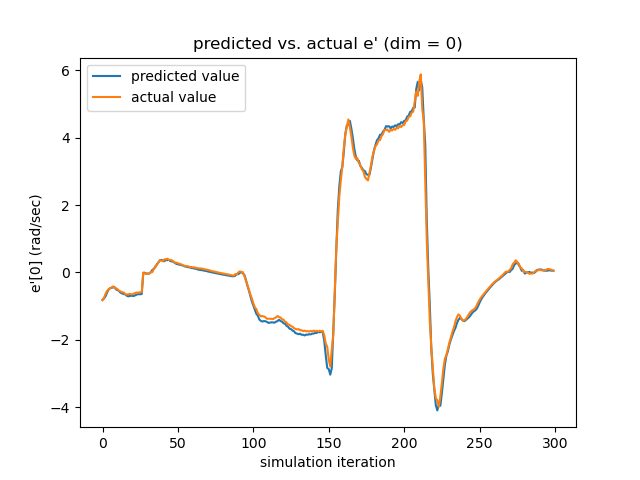}\hfill
    \includegraphics[width=0.33\textwidth]{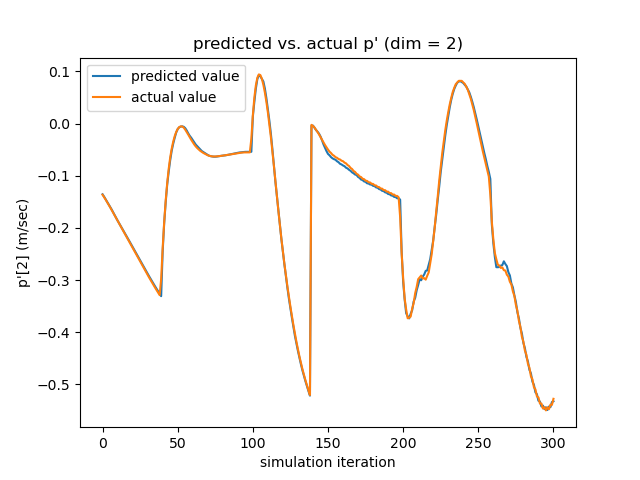}\hfill
    \includegraphics[width=0.33\textwidth]{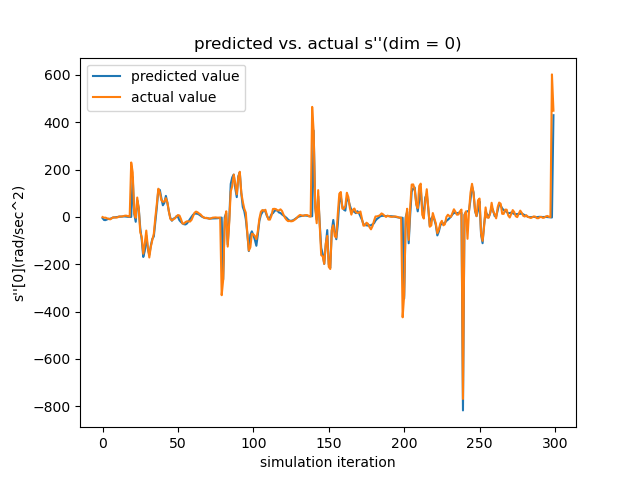}\hfill
    \includegraphics[width=0.33\textwidth]{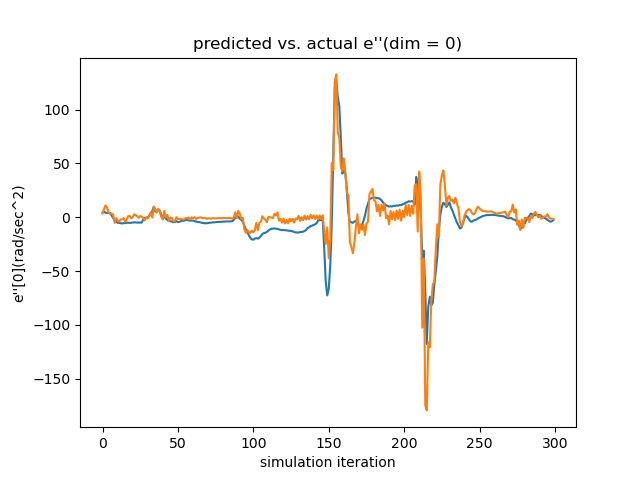}\hfill
    \includegraphics[width=0.33\textwidth]{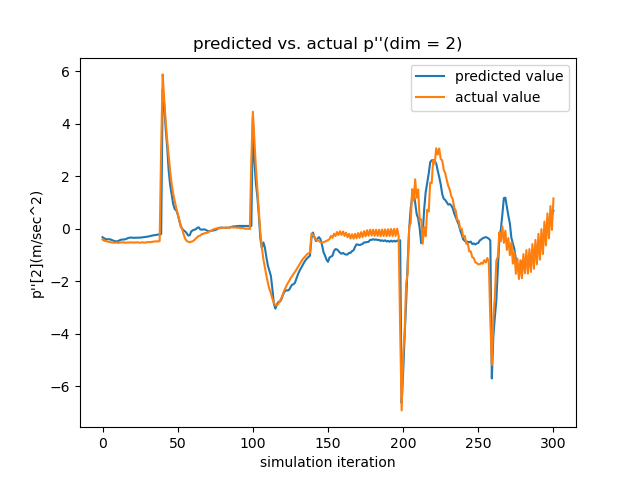}\hfill
    \caption{Testing results of the networks}\label{fig:networktesting}
\end{figure}
Section 4 of the paper presents the testing results of our trained neural dynamic systems. More testings are depicted in Figure. \ref{fig:networktesting} and Figure. \ref{fig:jacobtesting} with the same experimental setups. 


\section{Control}
\paragraph{LQR Overview}
\indent The Linear Quadratic Controller is a kind of full-state feedback controller, where the control of the system is based on the current state. Given a linear system in state-space form:

\begin{equation}
    \mathbf{\dot{x} = Ax + Bu}
\end{equation}
where \textbf{x} is the state vector, \textbf{u} is the control vector (in our case the thrusts for individual propellers), and \textbf{A} and \textbf{B} are matrices, we compute a control matrix \textbf{K} and combine that with the state by:

\begin{equation}
    \mathbf{u = -Kx}
\end{equation}

The way we obtain $\mathbf{K}$ is as follows: suppose we want to set both the state and the control to be \textbf{0}. We define cost matrices $\mathbf{Q}$ and $\mathbf{R}$ that penalizes $\mathbf{x}$ squared and $\mathbf{u}$ squared respectively, we desire to minimize the infinite horizon cost:

\begin{equation}
    \mathbf{\int_{0}^{\infty} [x^TQx + u^TRu] dt}
\end{equation}

which means our goal is to find the optimal cost-to-go function $
\mathbf{J^* = x^TSx}
$ that satisfies the Hamilton–Jacobi–Bellman (HJB) equation. Utilizing the convex nature of the problem, we know that the minimum occurs when the gradient is zero, so we have:
\begin{equation}
    \mathbf{\frac{\partial}{\partial u} = 2u^TR + 2x^TSB = 0}
\end{equation}
which yields the control policy:
\begin{equation}
    \mathbf{u = -[R^{-1}B^TS]x = -Kx}
\end{equation}
After transformation, we can find the value of $\mathbf{S}$ by solving the equation:
\begin{equation}
    \mathbf{0 = SA + A^TS - SBR^-1B^TS + Q}
\end{equation}

This is known as the \textbf{Algebraic Riccati Equation}, which can be solved by iterating backward in time.

\paragraph{Online Reinitialization Parameters}
Although our network eliminates the necessity for the extensive, empirical parameter tuning process, there are a few hyper-parameters that needs to be tuned for effective performance. We will present the exact value or the value range for these parameters in the table below.
\begin{center}
\begin{tabular}{ ll } 
 \hline
 Parameter type & Value/Value range \\ 
 \hline
  $\mathbf{Q}$ gain (related to $\mathbf{s}$) & 100 to 200\\
  \hline
  $\mathbf{Q}$ gain (related to $\mathbf{e}$) & 50 to 200\\
  \hline
  $\mathbf{Q}$ gain (related to $\mathbf{p}$) & 100 to 200\\
 \hline
 $\mathbf{R}$ gain & 2\\
 \hline
 $kp$  & $0.03$ \\
 \hline
 $kd$  & $0.0001$ \\
 \hline
 $n$  & $10$ \\
 \hline
\end{tabular}
\end{center}


\paragraph{Rigid LQR State Definition}
The state of a rigid object can be described by its position and rotation. Let $\mathbf{p}$ be the vector describing position, and let $\mathbf{e}$ be the vector describing rotation. And let $\mathbf{q} = 
\begin{pmatrix}
  \textbf p \\
  \textbf e \\
\end{pmatrix}$. 
Since the dynamics is second order, the state $\mathbf{x}$ will be defined as $\mathbf{x} =  
\begin{pmatrix}
  \textbf q \\
  \dot{\mathbf{q}} \\
\end{pmatrix}$. 
For the 3D case, $\mathbf{p} =
\begin{pmatrix}
  x \\
  y \\
  z \\
\end{pmatrix}$,  
\textbf e = $\begin{pmatrix}
  \phi \\
  \theta \\
  \psi \\
\end{pmatrix}$, where $x$, $y$, $z$ are the spacial coordinates, $\phi$, $\theta$, $\psi$ are the Euler angles. For the 2D case, $\mathbf{p} =  
\begin{pmatrix}
  x \\
  y \\
\end{pmatrix}$,  
$\mathbf{e} = \begin{pmatrix}
  \phi \\
\end{pmatrix}$, as the rotation in 2D can be described by a sole parameter.

\begin{center}
\begin{tabular}{ llp{8.5cm}}
 \hline
 $\mathbf{R}$ & SO(3) & Body-to-world rotation matrix\\ 
 \hline
  $\mathbf{r}$ & $R^{3}$ & Motor position in body frame \\
 \hline
  $\mathbf{d}$ & unit sphere & Motor orientation in body frame \\
 \hline
 $\mathbf{M_f}$ & $R^{3 \times n}$ & Mapping from thrusts to net force. The i-th column is $\mathbf{di}$. \\
 \hline
 $\mathbf{M_t}$ & $R^{3 \times n}$ & Mapping from thrusts to net torque. The i-th column is $bi \lambda i\mathbf{di} + ri \times \mathbf{di}$ \\  
 \hline
 $\mathbf{J}$ & $R^{3}$ & Inertia Tensor in Body Frame. Value is $\begin{bmatrix} I{xx} & I{xy} & I{xz} \\ I{yx} & I{yy} & I{yz} \\ I{zx} & I{zy} & I{zz} \end{bmatrix} $\\  
 \hline
 I{xx} & R & $\sum_{i}^{}mi*(yi^2+zi^2)$ \\
 I{yy} & R & $\sum_{i}^{}mi*(zi^2+xi^2)$ \\
 I{zz} & R & $\sum_{i}^{}mi*(xi^2+yi^2)$ \\
 I{xy} & R & $-\sum_{i}^{}mi*xi*yi$ \\
 I{xz} & R & $-\sum_{i}^{}mi*xi*zi$ \\
 I{yz} & R & $-\sum_{i}^{}mi*yi*zi$ \\
 \hline
 $\mathbf{L}$ & $R^{3}$ & Mapping from world frame angular velocity to body frame angular velocity, such that $\mathbf{\omega} = \mathbf{L\dot{e}}$. Value is 
 $\begin{bmatrix} 
    1 & 0 & $$-s(\theta)$$ \\
    1 & $$c(\theta)$$ & $$s(\phi)c(\theta)$$ \\
    1 & $$-s(\phi)$$ & $$c(\phi)c(\theta)$$ \\
\end{bmatrix} $ \\  
\hline
 $\mathbf{\dot{L}}$ & $R^{3}$ & Derivative of $\mathbf{L}$. Value is 
 $\begin{bmatrix} 
    0 & 0 & $$-c(\theta)\dot{\theta}$$ \\
    1 & $$-s(\phi)\dot{\phi}$$ & $$c(\phi)c(\theta)\dot{\phi}-s(\phi)s(\theta)\dot{\theta}$$ \\
    1 & $$-c(\phi)\dot{\phi}$$ & $$s(\phi)c(\theta)\dot{\phi}-c(\phi)s(\theta)\dot{\theta}$$ \\
\end{bmatrix} $ \\  
\hline
\end{tabular}
\end{center}

\paragraph{Dynamic Model}
Let $\mathbf{u}$ denote the drone's actuation, and $\mathbf{u} =
\begin{pmatrix}
  u_1 \\
  u_2 \\
  \vdots \\
  u_m \\
\end{pmatrix}$, where $m$ is the number of propellers and $u_i$ represents the thrust provided by each propeller. The dynamic model is a function \textbf{f} such that $\dot{\mathbf{x}} = \mathbf{f}(\mathbf{x}, \mathbf{u}) $.
In 3D, the dynamics of the drone will be directly derived from the Newton-Euler equations:
\begin{equation}
m\mathbf{\ddot{p}} = m\mathbf{g} + \mathbf{RM_fu}
\end{equation}
\begin{equation}
\mathbf{J(\dot{L}\dot{e} + L\ddot{e})} + (\mathbf{L\dot{e}}) \times \mathbf{JL\dot{e}}   = \mathbf{M_tu}
\end{equation}
with the variable definitions given in the table below.
For the 2D case, these equations simplify to
\begin{equation}
m\mathbf{\ddot{p}} = m\mathbf{g} + \mathbf{RM_fu}
\end{equation}
\begin{equation}
\mathbf{J\ddot{e}}=\mathbf{M_tu}
\end{equation}
\paragraph{Manipulator Form}
Follow the formulation purposed in ~\citeS{2russnote}, we will reorganize these equations into the Manipulator Form, whose template is as follows:
\begin{equation}
    \mathbf{H(q)\ddot{q} + C(q,\dot{q})\dot{q} + G(q) = B(q)u},
\end{equation} Consequently,
\begin{equation}
    \mathbf{\ddot{q} = H^{-1}(B(q)u - C(q,\dot{q})\dot{q} - G(q))}
\end{equation}
This allows us to write: 
\begin{equation} 
\mathbf{f(x,u) = \dot{x} = } 
\begin{bmatrix} 
\mathbf{\dot{q}}\\
\mathbf{\ddot{q}}\\
\end{bmatrix}
 = 
\begin{bmatrix} 
\mathbf{\dot{q}}\\
\mathbf{H^{-1}(B(q)u - C(q,\dot{q})\dot{q} - G(q))}\\
\end{bmatrix}
\end{equation}
For the 3D case, reorganizing the dynamics equations yields
$\mathbf{H} = \begin{bmatrix} 
m\mathbf{I_3} & \mathbf{O} \\
\mathbf{O} & \mathbf{JL} \\
\end{bmatrix}
$,
$ \mathbf{C} = 
\begin{bmatrix} 
\mathbf{O} & \mathbf{O} \\
\mathbf{O} & \mathbf{J\dot{L} + L\dot{e} \times JL} \\
\end{bmatrix}
$, 
$
\mathbf{G} = 
\begin{bmatrix} 
-m\mathbf{g} \\
\mathbf{O}\\
\end{bmatrix}
$, 
$
\mathbf{B} = 
\begin{bmatrix} 
\mathbf{RM_f} \\
\mathbf{M_t}\\
\end{bmatrix}
$.
\\For the 2D case, we have 
$\mathbf{H} = \begin{bmatrix} 
m\mathbf{I_2} & \mathbf{O} \\
\mathbf{O} & \mathbf{J} \\
\end{bmatrix}
$,
$\mathbf{C} = 
\begin{bmatrix} 
\mathbf{O}\\
\end{bmatrix}
$, 
$
\mathbf{G} = 
\begin{bmatrix} 
-m\mathbf{g} \\
\mathbf{O}\\
\end{bmatrix}
$, 
$
\mathbf{B} = 
\begin{bmatrix} 
\mathbf{RM_f} \\
\mathbf{M_t}\\
\end{bmatrix}
$.

\paragraph{Linearization via Taylor Expansion}
\indent Since the function $\mathbf{f(x,u)}$ described above is a non-linear model, we will linearize it by taking the first order Taylor Expansion around an operating point ($\mathbf{x^*, u^*}$) such that $\mathbf{f(x^*, u^*) = 0}$. For $\mathbf{x}$ close enough to $\mathbf{x^*}$, we have: 
\begin{equation}
\begin{split}
\mathbf{f(x-x^*)}
& \approx (\frac{\partial f}{\partial x}|{\substack{\mathbf{x=x^*, u=u^*}}})\mathbf{(x-x^*)} + \mathbf{(\frac{\partial f}{\partial u}|\substack{\mathbf{x=x^*, u=u^*}})(u-u^*)}\\ 
& = \mathbf{A_{lin}}\mathbf{(x-x^*)} + \mathbf{B_{lin}}\mathbf{(u-u^*)}
\end{split}
\end{equation}

Since we know that: 
\begin{equation} \mathbf{f(x,u)= }
\begin{bmatrix} 
\mathbf{\dot{q}}\\
\mathbf{H^{-1}(B(q)u - C(q,\dot{q})\dot{q} - G(q))}\\
\end{bmatrix}
, 
\mathbf{x} =
\begin{pmatrix}
  \textbf q \\
  \dot{\mathbf{q}} \\
\end{pmatrix},
\end{equation}
$\mathbf{A_{lin}} = \mathbf{\frac{\partial f}{\partial x}}$ can be represented by the block matrix:
\begin{equation}
\begin{bmatrix} 
\mathbf{\frac{\partial \dot{q}}{\partial q}} & \mathbf{\frac{\partial \dot{q}}{\partial \dot{q}}} \\
\mathbf{\frac{\partial H^{-1}(B(q)u - C(q,\dot{q})\dot{q} - G(q))}{\partial q}} & \mathbf{\frac{\partial H^{-1}(B(q)u - C(q,\dot{q})\dot{q} - G(q))}{\partial \dot{q}}} \\
\end{bmatrix} = \begin{bmatrix} 
\mathbf{T1} & \mathbf{T2} \\
\mathbf{T3} & \mathbf{T4} \\
\end{bmatrix}
\end{equation}

It can be seen trivially that $\mathbf{T1} = \mathbf{\frac{\partial \dot{q}}{\partial q}} = \mathbf{O}$ and $\mathbf{T2} = \mathbf{\frac{\partial \dot{q}}{\partial \dot{q}}} = \mathbf{I_3}$.

For $\mathbf{T_3} = \mathbf{\frac{\partial H^{-1}(B(q)u - C(q,\dot{q})\dot{q} - G(q))}{\partial q}}$, by the Product Rule we know, 
\begin{equation}
\mathbf{T_3} = \mathbf{\frac{\partial H^{-1}}{\partial q}} \mathbf{(B(q)u - C(q,\dot{q})\dot{q} - G(q))} + \mathbf{H^{-1} \mathbf{\frac{\partial (B(q)u - C(q,\dot{q})\dot{q} - G(q))}{\partial q}}}
\end{equation}
Since we defined $\mathbf{x^*, u^*}$ to be such that $\mathbf{f(x^*, u^*) = 0}$, then $\mathbf{H^{-1}(B(q)u - C(q,\dot{q})\dot{q} - G(q)) = 0}$ at $\mathbf{(x^*, u^*)}$. Since we know $\mathbf{H^{-1}}$ is non-zero, then $\mathbf{B(q)u - C(q,\dot{q})\dot{q} - G(q) = 0}$. Besides, since we have $\mathbf{f(x^*, u^*) = 0}$, we have $\mathbf{\dot{q} = 0}$, then 
$\mathbf{\frac{\partial C(q,\dot{q})\dot{q}}{\partial q}} = \mathbf{\frac{\partial C(q,\dot{q})}{\partial q} \dot{q}} + \mathbf{C(q,\dot{q})}\mathbf{\frac{\partial\dot{q}}{\partial q}} = \mathbf{0} + \mathbf{0} =  \mathbf{0}. $
Also, since $\mathbf{G(q)} = 
\begin{bmatrix} 
-m\mathbf{g} \\
\mathbf{O}\\
\end{bmatrix}
$, and has nothing to do with $\mathbf{q}$, 
$\mathbf{\frac{\partial G(q)}{\partial q}} = \mathbf{0}$. So we can conclude that:
\begin{equation}
\mathbf{T_3} = \mathbf{H^{-1}} \mathbf{\frac{\partial B(q)u}{\partial q}} = \mathbf{H^{-1}} (\mathbf{\frac{\partial (B(q)}{\partial q}u + B(q)\frac{\partial u}{\partial q}}) = \mathbf{H^{-1}} \mathbf{\frac{\partial (B(q)}{\partial q}u}. 
\end{equation}
Since
\begin{equation}
 \mathbf{B} = 
\begin{bmatrix} 
\mathbf{RM_f} \\
\mathbf{M_t}\\
\end{bmatrix}
 = 
\begin{bmatrix}
\begin{pmatrix}
\mathbf{(RM_f)_1} \\
\mathbf{(M_t)_1} \\
\end{pmatrix} & 
\begin{pmatrix}
\mathbf{(RM_f)_2} \\
\mathbf{(M_t)_2} \\
\end{pmatrix} & 
\begin{pmatrix}
\mathbf{(RM_f)_3} \\
\mathbf{(M_t)_3} \\
\end{pmatrix} & 
\begin{pmatrix}
\mathbf{(RM_f)_4} \\
\mathbf{(M_t)_4} \\
\end{pmatrix}\\
\end{bmatrix}
\end{equation}
\begin{equation}
\mathbf{\frac{\partial B}{\partial q}} = 
\begin{bmatrix}
\frac{\partial\begin{pmatrix}
\mathbf{(RM_f)_1} \\
\mathbf{(M_t)_1} \\
\end{pmatrix}}{\mathbf{\partial q}} & 
\frac{\partial\begin{pmatrix}
\mathbf{(RM_f)_2} \\
\mathbf{(M_t)_2} \\
\end{pmatrix}}{\mathbf{\partial q}}  & 
\frac{\partial\begin{pmatrix}
\mathbf{(RM_f)_3} \\
\mathbf{(M_t)_3} \\
\end{pmatrix}}{\mathbf{\partial q}}  & 
\frac{\partial\begin{pmatrix}
\mathbf{(RM_f)_4} \\
\mathbf{(M_t)_4} \\
\end{pmatrix}}{\mathbf{\partial q}} 
\end{bmatrix}.
\end{equation}
So,
\begin{equation}
\mathbf{\frac{\partial B}{\partial q}} \mathbf{u} =
\sum_{i}^{} \frac{\partial\begin{pmatrix}
\mathbf{(RM_f)_i} \\
\mathbf{(M_t)_i} \\
\end{pmatrix}}{\mathbf{\partial q}} * u_i,
\end{equation}
where:
\begin{equation}
\begin{split}
\frac{\partial\begin{pmatrix}
\mathbf{(RM_f)_i} \\
\mathbf{(M_t)_i} \\
\end{pmatrix}}{\mathbf{\partial q}}
& = \begin{bmatrix}
\frac{\partial\mathbf{(RM_f)_i}}{\partial\mathbf{p}} & \frac{\partial\mathbf{(RM_f)_i}}{\partial\mathbf{e}} \\
\frac{\partial\mathbf{(M_t)_i}}{\partial\mathbf{p}} & \frac{\partial\mathbf{(M_t)_i}}{\partial\mathbf{e}} \\
\end{bmatrix} = 
\begin{bmatrix}
\mathbf{O_{3\times3}} & \frac{\partial\mathbf{R}}{\partial\mathbf{e}} * \mathbf{(M_f)_i} \\
\mathbf{O_{3\times3}} & \mathbf{O_{3\times3}} \\
\end{bmatrix}
\\
& = 
\begin{bmatrix}
\mathbf{O_{3\times3}} & \begin{bmatrix}
\frac{\partial\mathbf{R}}{\partial\mathbf{\phi}} * \mathbf{(M_f)_i} 
& \frac{\partial\mathbf{R}}{\partial\mathbf{\theta}} * \mathbf{(M_f)_i} 
& \frac{\partial\mathbf{R}}{\partial\mathbf{\psi}} * \mathbf{(M_f)_i} 
\end{bmatrix}\\
\mathbf{O_{3\times3}} & \mathbf{O_{3\times3}} \\
\end{bmatrix}.
\end{split}
\end{equation}
Finally, 
$$\mathbf{T_4} = \mathbf{\frac{\partial H^{-1}}{\partial \dot{q}}} \mathbf{(B(q)u - C(q,\dot{q})\dot{q} - G(q))} + \mathbf{H^{-1} \mathbf{\frac{\partial (B(q)u - C(q,\dot{q})\dot{q} - G(q))}{\partial \dot{q}}}} = \mathbf{-H^{-1}C},$$ since $\mathbf{B(q)u - C(q,\dot{q})\dot{q} - G(q)=0}$ and with $\mathbf{\dot{q} = 0}$, $\mathbf{\frac{\partial (B(q)u - C(q,\dot{q})\dot{q} - G(q))}{\partial \dot{q}}} = \mathbf{\frac{\partial (- C(q,\dot{q})\dot{q})}{\partial \dot{q}}} = \mathbf{- C(q,\dot{q}))\frac{\partial\dot{q}}{\partial \dot{q}} = - C(q,\dot{q})}$.

\indent So to sum up:

\begin{equation}
\mathbf{A_{lin}} = 
\begin{bmatrix}
\mathbf{O_{6\times6}} & \mathbf{I_{6\times6}} \\
\mathbf{H^{-1}} * \sum_{i}^{} \begin{bmatrix}
\mathbf{O_{3\times3}} & \begin{bmatrix}
\frac{\partial\mathbf{R}}{\partial\mathbf{\phi}} * \mathbf{(M_f)_i} 
& \frac{\partial\mathbf{R}}{\partial\mathbf{\theta}} * \mathbf{(M_f)_i} 
& \frac{\partial\mathbf{R}}{\partial\mathbf{\psi}} * \mathbf{(M_f)_i} 
\end{bmatrix}\\
\mathbf{O_{3\times3}} & \mathbf{O_{3\times3}} \\
\end{bmatrix} * u_i& \mathbf{-H^{-1}C} \\
\end{bmatrix}
\end{equation}
\\For $\mathbf{B_{lin}}$, we have:
\begin{equation}
\mathbf{B_{lin}} = \mathbf{\frac{\partial f}{\partial u}} = 
\begin{bmatrix} 
\mathbf{\frac{\partial \dot{q}}{\partial u}} \\
\mathbf{\frac{\partial H^{-1}(B(q)u - C(q,\dot{q})\dot{q} - G(q))}{\partial u}}\\
\end{bmatrix}
\end{equation}.

It is clear to see that $\mathbf{\frac{\partial \dot{q}}{\partial u}} = \mathbf{0}$, and $\mathbf{\frac{\partial H^{-1}(B(q)u - C(q,\dot{q})\dot{q} - G(q))}{\partial u}} = \mathbf{\frac{\partial H^{-1}(B(q)u)}{\partial u}} = \mathbf{H^{-1} B(q)}\mathbf{\frac{\partial u}{\partial u}} = \mathbf{H^{-1} B(q)}$, (remember than none of \textbf{G, B, C, H} is related to \textbf{u}).
So we have 
\begin{equation}
\mathbf{B_{lin}} = 
\begin{bmatrix} 
\mathbf{O_{6\times k}} \\
\mathbf{H^{-1} B}\\
\end{bmatrix}
\end{equation}

For the 2D case $\mathbf{A_{lin}}$ and $\mathbf{B_{lin}}$ are simplified to become:
\begin{equation}
\begin{split}
\mathbf{A_{lin}} & = \mathbf{\frac{\partial f}{\partial x}} = 
\begin{bmatrix} 
\mathbf{\frac{\partial \dot{q}}{\partial q}} & \mathbf{\frac{\partial \dot{q}}{\partial \dot{q}}} \\
\mathbf{\frac{\partial H^{-1}(B(q)u - C(q,\dot{q})\dot{q} - G(q))}{\partial q}} & \mathbf{\frac{\partial H^{-1}(B(q)u - C(q,\dot{q})\dot{q} - G(q))}{\partial \dot{q}}} \\
\end{bmatrix}
\\
& = 
\begin{bmatrix}
\mathbf{O_{3\times3}} & \mathbf{I_{3\times3}} \\
\mathbf{H^{-1}} * \sum_{i}^{} \begin{bmatrix}
\mathbf{O_{2\times2}} & \frac{\partial\mathbf{R}}{\partial\mathbf{\phi}} * \mathbf{(M_f)_i}\\
\mathbf{O_{1\times2}} & \mathbf{O_{1\times1}} \\
\end{bmatrix} * u_i& \mathbf{O_{3\times3}} \\
\end{bmatrix}
\end{split}
\end{equation}

\begin{equation}
\mathbf{B_{lin}} = 
\begin{bmatrix} 
\mathbf{\frac{\partial \dot{q}}{\partial u}} \\
\mathbf{\frac{\partial H^{-1}(B(q)u - C(q,\dot{q})\dot{q} - G(q))}{\partial u}}\\
\end{bmatrix} =
\begin{bmatrix} 
\mathbf{O_{3\times k}} \\
\mathbf{H^{-1} B}\\
\end{bmatrix}
\end{equation}
The matrices $\mathbf{A_{lin}}$ and $\mathbf{B_{lin}}$ will then be optimized by the LQR to yield the control matrix $\mathbf{K}$. For the geometry-updating LQR, the quantities that are updated each time are $\mathbf{M_{f}}$, $\mathbf{M_{t}}$ and $\mathbf{J}$, which are all the time-varying values in the above derivation. Besides, the fixed-point is also recalculated.
\paragraph{Fixed Point}
Assuming $\mathbf{(x^{*}, u^{*})}$ satisfies $\mathbf{f(x^{*}, u^{*}) = 0}$, we have:
\begin{equation}
\mathbf{RM_{f}u^{*}} = -m\mathbf{g}
\end{equation}
\begin{equation}
\mathbf{M_{t}u^{*} = 0}
\end{equation}
for torque and force balance respectively.
Given certain $\mathbf{M_{f}}$, $\mathbf{M_{t}}$, we will first solve the equation:
\begin{equation}
\begin{bmatrix}
  \mathbf {1} \\
  \mathbf{M_{t}} \\
\end{bmatrix} \mathbf{u} 
= \begin{pmatrix}
  -||m\mathbf{g}|| \\
  \mathbf{0} \\
\end{pmatrix}
\end{equation}
to satisfy the torque balance. Then we will rotate the reference frame so that the direction of the combined thrust aligns with the $Y$-axis. The rotation axis is calculated by: 
\begin{equation}
\mathbf{v} = \mathbf{M_{f}u^{*}\times}(-m\mathbf{g})
\end{equation}
\begin{equation}
\mathbf{\hat{v}} = \frac{\mathbf{v}}{||\mathbf{v}||}
\end{equation}
The angle is calculated by:
\begin{equation}
\alpha = \frac{-m\mathbf{g}^{T}\mathbf{M_{f}u^{*}}}{||\mathbf{M_{f}u^{*}}||||m\mathbf{g}||}
\end{equation}
After the rotation in axis-angle form is calculated, the fixed-point Euler angles would be extracted to form the $\mathbf{e^{*}}$ part of $\mathbf{x^{*}}$.

\begin{figure}[hbt!]
    \centering
    \includegraphics[width=0.99\textwidth]{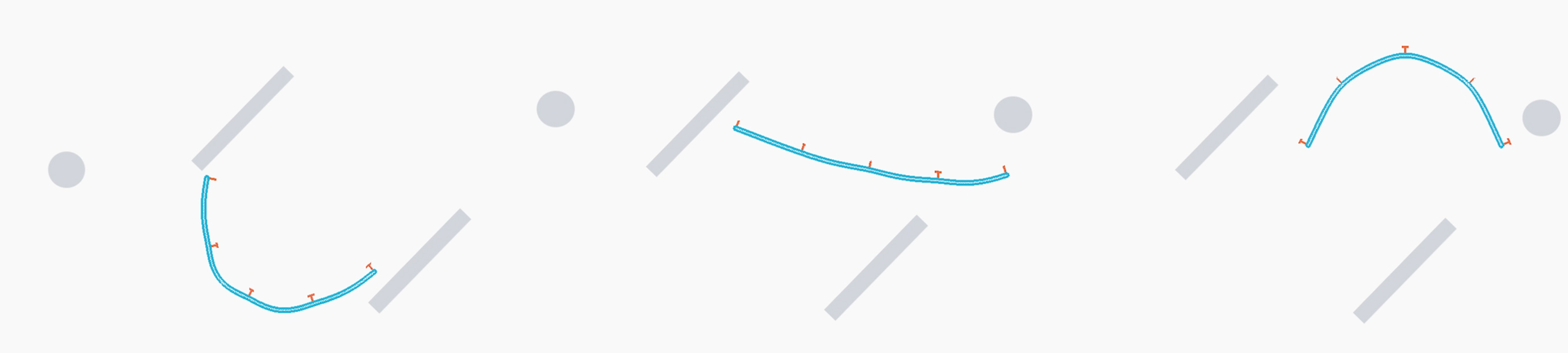}\hfill
    \includegraphics[width=0.99\textwidth]{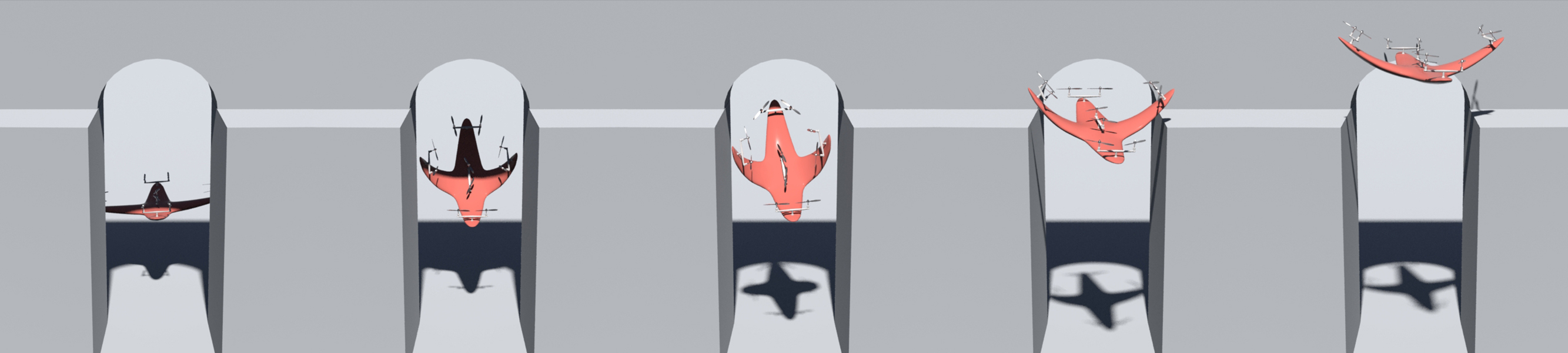}\hfill
    \includegraphics[width=0.99\textwidth]{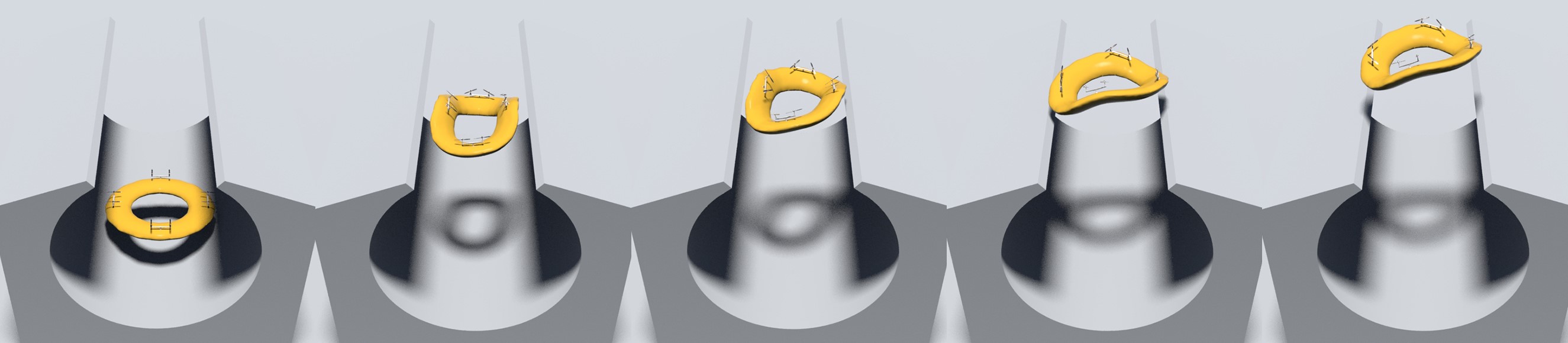}\hfill
    \includegraphics[width=0.99\textwidth]{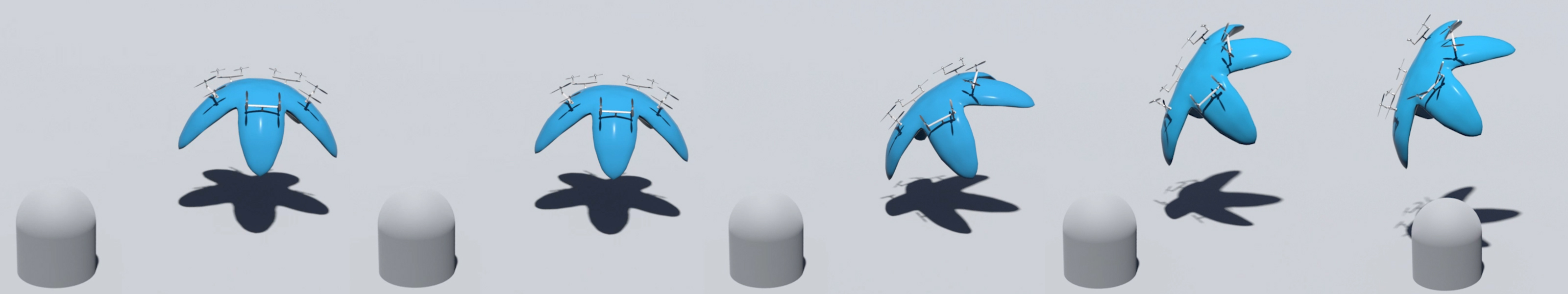}\hfill
    \includegraphics[width=0.99\textwidth]{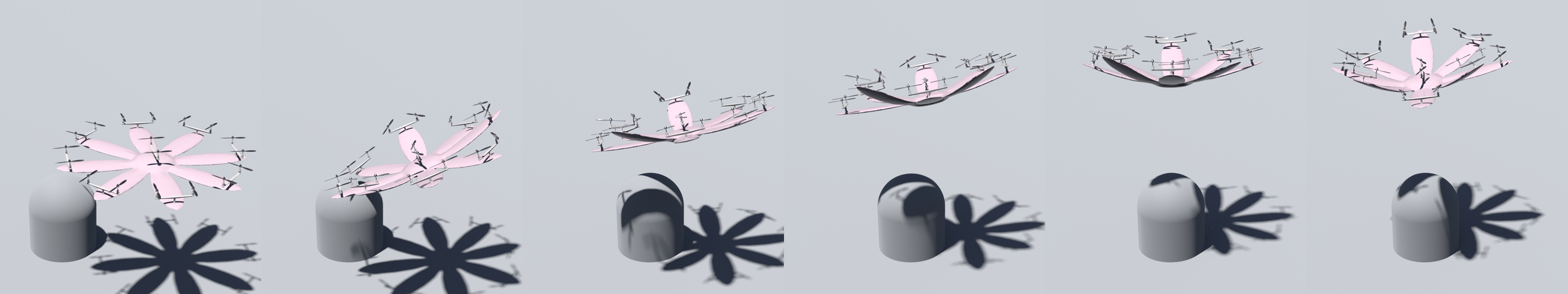}\hfill
    \includegraphics[width=0.99\textwidth]{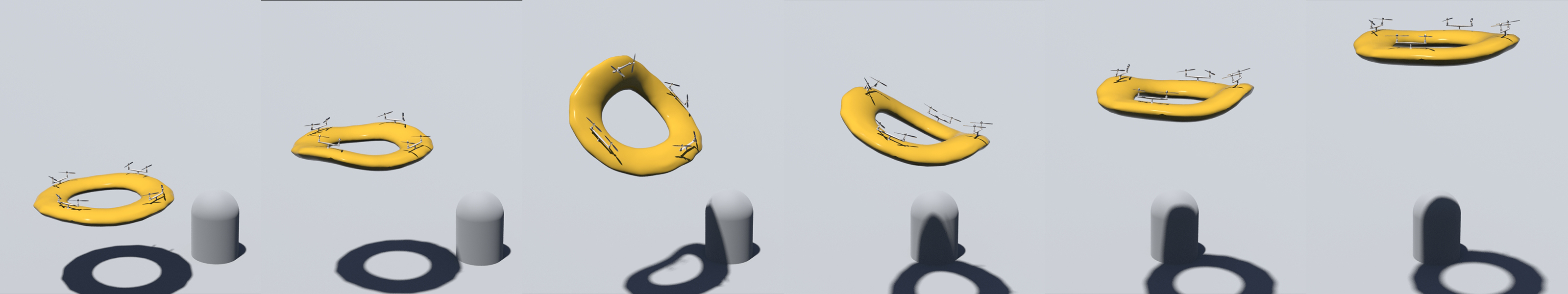}\hfill
    \includegraphics[width=0.99\textwidth]{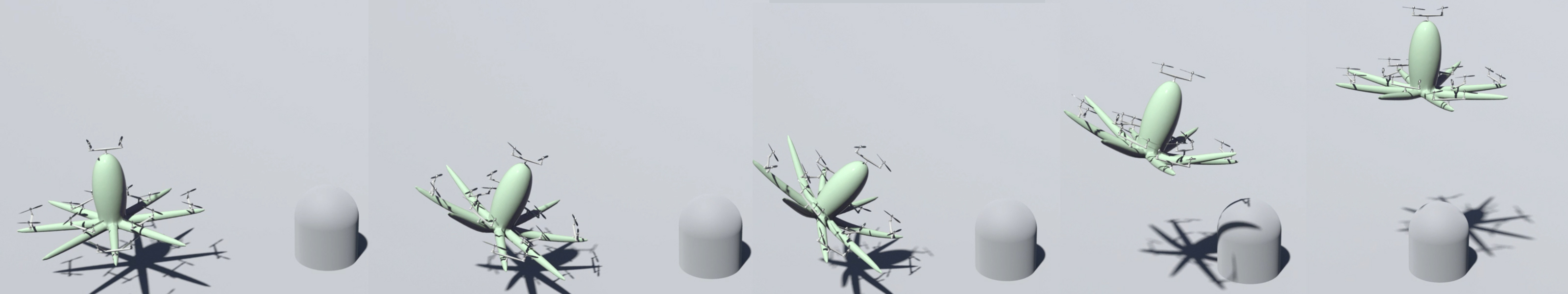}\hfill
    \caption{Visualization of more test results; Top 3: Obstacle avoidance animation; Bottom: Locomotion animation}\label{fig:results}
\end{figure}

\section{Toward a Real Soft Drone}

Although we carried out the experiments purely in numerical simulation environments, we designed our approach with its real-life feasibility in mind, and our method is intrinsically suitable for real-world deployment. 
First, our perception of the soft drone is explicitly sensor-based. Unlike many other works that deal with soft-robot controls like ~\citeS{2barbivc2008reallqrsoft}~\citeS{2spielberg2019learning} which observe the full state (particle positions) and apply model reduction techniques to synthesize the state, we resist this unrealistic assumption, and throughout our pipeline, the interfacing between the simulator and the training/controlling modules is strictly limited to the sensor measurements. In this sense, we observe the simulation environment in the same limited fashion as we observe the real world, so that no unfair advantage is taken. We expect the rest of the pipeline to work exactly the same if we swap the simulator with the real-world environment, since the interfacing will not be changed. Secondly, as we have mentioned in the paper, in designing the sensing scheme, the only sensors we used are Inertial Measurement Units (IMU), which are basic and accessible tools used everywhere for rigid drones. No other sensor types, such as bending, thermal or fluidic sensors are used. This simplistic approach allows us to conveniently fabricate these soft drones by implanting the IMU microprocessors at the surface, without having to cut open the drone's body or insert extra measurement devices. Basically, to fabricate an actual soft drone, we just need to cut out the desired shape from solid materials (if not with 3D printing techniques), implant the IMUs at the surface, set up their connection to a central onboard processor using WIFI, and flash the trained neural network and controlling script onto the hardware. Thirdly, the computational efficiency of our algorithm allows it to be handled by on-board processors. In the testing case, our relinearization is done at $10Hz$, and can be relaxed to $20Hz$ for the more stable geometries. There have been previous works conducted that performs LQR recalculation~\citeS{2foehn2018onboardLQR} and network-based control loop~\citeS{2kaufmann2018deep} at $10Hz$ using onboard computers. With further code optimization, we believe that our current system can be implemented fully onboard. Lastly, we simulate the soft body using a co-rotated elastic Finite Element simulator, which is known for providing physically realistic behaviors and is commonly used in engineering design, with noise and time delay applied. As a result, our success in this simulation testing environment is meaningful for its future adaptation to real-world scenarios.

\clearpage
\bibliographystyleS{plain}
\bibliographyS{example}

\include{pythonlisting}

\title{Soft Drones Flying: Control of Deformable Multicopters via Dynamics Identification}
\title{
Soft Multicopter Control using\\
Neural Dynamics Identification \\
Supplementary
}

\end{document}